\documentclass[sigconf]{acmart}
\usepackage{newcommand}
\usepackage{amsbsy}
\usepackage{multicol,graphicx,mathrsfs,pifont,amscd,latexsym,color,fancyhdr,url,longtable, pifont, subfigure,epstopdf,setspace,multirow,colortbl,tabularx, booktabs, bbm,listings, balance,color,indentfirst,algorithmic,algorithm, subfigure}
\usepackage{paralist}
\usepackage{amsmath}
\usepackage{stackengine}
\usepackage{relsize}
\usepackage{bbold}
\usepackage{amsfonts}
\usepackage{xspace}
\usepackage{amssymb}
\usepackage{multirow}

\usepackage{amsthm}

\theoremstyle{definition}

\newcommand{\method}{GPT-GNN\xspace}

\newcommand{\zn}[1]{\hide{{\color{red}#1 }}}
\newcommand{\yd}[1]{\hide{{\color{red}#1 }}}
\newcommand{\kw}[1]{\hide{{\color{blue}[kw: #1 ]}}}
\newcommand{\YS}[1]{\hide{{\color{purple}[YS: #1 ]}}}

\newcommand{\hide}[1]{} 
\newcommand{\vpara}[1]{\vspace{0.05in}\noindent\textbf{#1 }}

\copyrightyear{2020}
\acmYear{2020}
\setcopyright{rightsretained}
\acmConference[KDD '20]{Proceedings of the 26th ACM SIGKDD Conference on Knowledge Discovery and Data Mining}{August 23--27, 2020}{Virtual Event, CA, USA}
\acmBooktitle{Proceedings of the 26th ACM SIGKDD Conference on Knowledge Discovery and Data Mining (KDD '20), August 23--27, 2020, Virtual Event, CA, USA}
\acmDOI{10.1145/3394486.3403237}
\acmISBN{978-1-4503-7998-4/20/08}

\begin{document}
\fancyhead{}

\title{\method: Generative Pre-Training of Graph Neural Networks}

\author{Ziniu Hu}
\affiliation{%
  \institution{University of California, Los Angeles}
}
\email{bull@cs.ucla.edu}

\author{Yuxiao Dong}
\affiliation{%
  \institution{Microsoft Research, Redmond}
}
\email{yuxdong@microsoft.com}

\author{Kuansan Wang}
\affiliation{%
  \institution{Microsoft Research, Redmond}
}
\email{kuansanw@microsoft.com}

\author{Kai-Wei Chang}
\affiliation{%
  \institution{University of California, Los Angeles}
}
\email{kwchang@cs.ucla.edu}

\author{Yizhou Sun}
\affiliation{%
  \institution{University of California, Los Angeles}
}
\email{yzsun@cs.ucla.edu}

\begin{abstract}
Graph neural networks (GNNs) have been demonstrated to be powerful in modeling graph-structured data. 
However, training GNNs usually requires abundant task-specific labeled data, which is often arduously expensive to obtain. 
One effective way to reduce the labeling effort is to pre-train an expressive GNN model on unlabeled data with self-supervision and then transfer the learned model to downstream tasks with only a few labels. 
In this paper, we present the \method\footnote{The code and pre-trained models are available at \url{https://github.com/acbull/GPT-GNN}.} framework to initialize GNNs by generative pre-training. 
\method introduces a self-supervised attributed graph generation task to pre-train a GNN so that it can capture the structural and semantic properties of the graph. 
We factorize the likelihood of the graph generation into two components: 1) Attribute Generation and 2) Edge Generation. 
By modeling both components, \method\ captures the inherent dependency between node attributes and graph structure during the generative process. 
Comprehensive experiments on the billion-scale Open Academic Graph and Amazon recommendation data demonstrate that  \method\ significantly outperforms state-of-the-art GNN models without pre-training by up to 9.1\% across various downstream tasks.

\end{abstract}





\maketitle

\section{Introduction}\label{sec:introduction}


The breakthroughs in graph neural networks (GNNs) have revolutionized graph mining from structural feature engineering to representation learning~\cite{bruna2013spectral, gilmer2017neural,gcn}. 
Recent GNN developments have been demonstrated to benefit various graph applications and network tasks, such as semi-supervised node classification~\cite{gcn}, recommendation systems~\cite{DBLP:conf/kdd/YingHCEHL18}, and knowledge graph inference~\cite{DBLP:conf/esws/SchlichtkrullKB18}.

Commonly, GNNs take a graph with attributes as input and apply convolutional filters to generate node-level representations layer by layer. 
Often, a GNN model is trained with supervised information in an end-to-end manner for one task on the input graph. 
That said, for different tasks on the same graph, it is required to have enough and different sets of labeled data to train dedicated GNNs corresponding to each task.  
Usually, it is arduously expensive and sometimes infeasible to access sufficient labeled data for those tasks, particularly for large-scale graphs. 
Take, for example, the author disambiguation task in academic graphs~\cite{tang2008arnetminer}, it has still faced the challenge of the lack of ground-truth to date.

Similar issues had also been experienced in natural language processing (NLP). 
Recent advances in NLP address them by training a model from a large unlabeled corpus and transferring the learned model to downstream tasks with only a few labels---the idea of pre-training. 
For example, the pre-trained BERT language model~\cite{DBLP:journals/corr/abs-1810-04805} is able to learn expressive contextualized word representations by reconstructing the input text---next sentence and masked language predictions, and thus it can significantly improve the performance of various downstream tasks. 
Additionally, similar observations have also been demonstrated in computer vision~\cite{DBLP:journals/corr/abs-1807-03748, he2019momentum, DBLP:journals/corr/abs-2002-05709}.

Inspired by these developments, we propose to pre-train graph neural networks for graph mining. The goal of the pre-training is to empower GNNs to capture the structural and semantic properties of a input graph, so that it can easily generalize to any downstream tasks with a few fine-tuning steps on the graphs within the same domain. To achieve this goal, we propose to model the graph distribution by learning to reconstruct the input attributed graph.

To pre-train GNNs based on graph reconstruction, one straightforward option could be to directly adopt the neural graph generation techniques~\cite{ DBLP:journals/corr/KipfW16a, DBLP:conf/icml/YouYRHL18, DBLP:conf/nips/LiaoLSWHDUZ19}. 
However, they are not suitable for pre-training GNNs by design. 
First, most of them focus only on generating graph structure without attributes, which does not capture the underlying patterns between node attributes and graph structure---the core of convolutional aggregation in GNNs. Second, they are designed to handle small graphs to date, limiting their potential to pre-train on large-scale graphs. 

\begin{figure}[t!]
    \centering
    \includegraphics[width=0.475\textwidth]{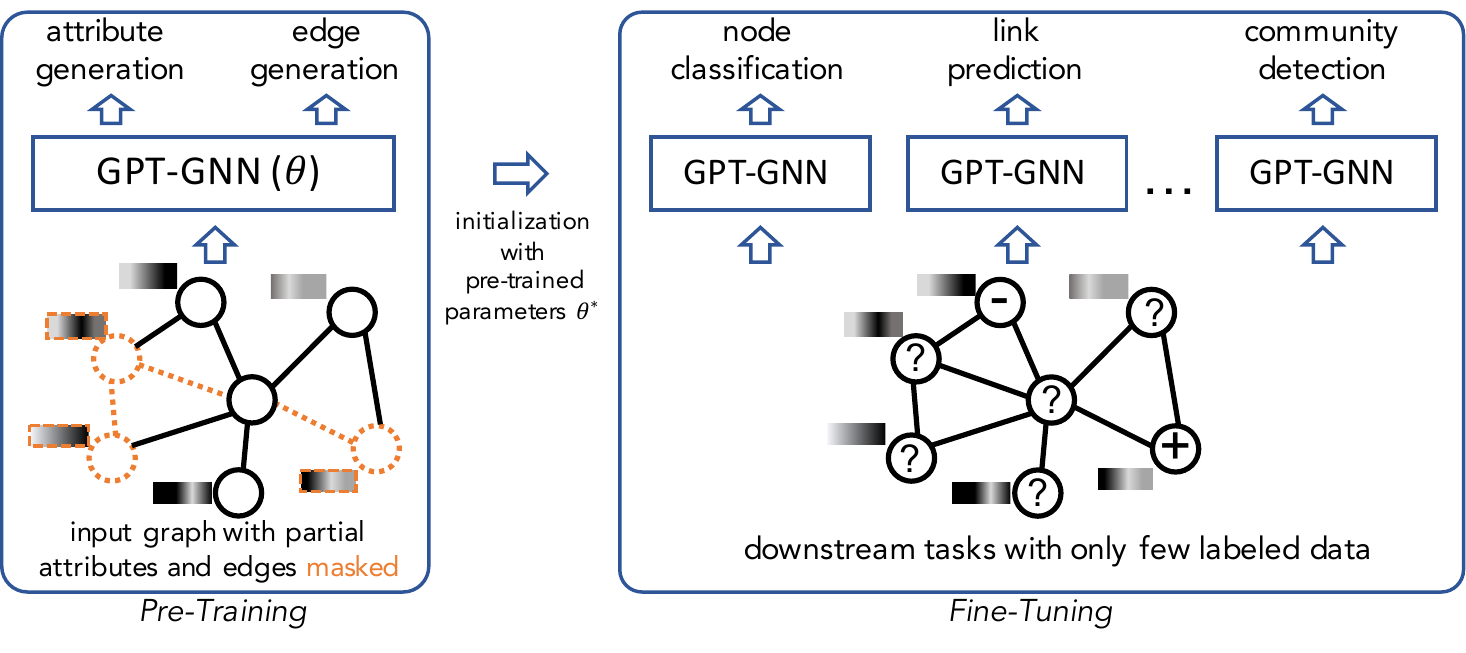}
    \caption{The pre-training and fine-tuning flow of \method: First, a GNN is pre-trained with the self-supervised learning task---attribute and structure generations. Second, the pre-trained model and its parameters are then used to initialize models for downstream tasks on the input graph or graphs of the same domain.}
    \label{fig:arch}
\end{figure}

\vpara{Contributions.}
In this work, we design a self-supervised attributed graph generation task for GNN pre-training, with which both the structure and attributes of the graph are modeled. Based on this task, we present the \method\ framework for generative pre-training of graph neural networks (Cf. Figure \ref{fig:arch}). 
The pre-trained GNN on the input graph can be then used as the initialization of models for 
different downstream tasks on the same type of graphs. 
Specifically, our contributions are illustrated below. 

First, we design an attributed graph generation task to model both node attributes and graph structure. We decompose the graph generation objective into two components: Attribute Generation and Edge Generation, whose joint optimization is equivalent to maximizing the probability likelihood of the whole attributed graph. 
In doing this, the pre-trained model can capture  the inherent dependency between node attributes and graph structure.

Second, we propose an efficient framework \method to conduct generative pre-training with the aforementioned task. \method can calculate the attribute and edge generation losses of each node simultaneously, and thus only need to run the GNN once for the graph. 
Additionally, \method can handle large-scale graphs with sub-graph sampling and mitigate the inaccurate loss brought by negative sampling with an adaptive embedding queue.

Finally, we pre-train GNNs on two large-scale graphs---the Open Academic Graph (OAG) of 179 million nodes \& 2 billion edges and Amazon recommendation data of 113 million nodes. 
Extensive experiments show that the \method\ pre-training framework can significantly benefit various downstream tasks. 
For example, by applying the pre-trained model on OAG, the node classification and link prediction performance is on average lifted by 9.1\% over the state-of-the-art GNN models without pre-training. 
In addition, we show that \method\ 
can consistently improve the performance of different base GNNs under various settings.

\section{Preliminaries and Related Work}\label{sec:related}

The goal of pre-training is to allow a model (usually neural networks) to initialize its parameters with pre-trained weights. 
In this way, the model can leverage the commonality between the pre-training and  downstream tasks. Recently pre-training has shown superiority in boosting the performance of many downstream applications in computer vision and natural language processing. In the following, we first introduce the preliminaries about GNNs and then review pre-training approaches in graphs and other domains.

\subsection{Preliminaries of Graph Neural Networks}

Recent years have witnessed the success of GNNs for modeling graph data~\cite{gcn,gat,graphsage,hgt}. 
A GNN can be regarded as using the input graph structure as the computation graph for message passing~\cite{gilmer2017neural}, during which the local neighborhood information is aggregated to get a more contextual representation. Formally, suppose $H^{(l)}_t$ is the node representation of node $t$ at the $(l)$-th GNN layer, the update procedure from the $(l$-$1)$-th layer to the $(l)$-th layer is:
\begin{align}
H^{(l)}_t \gets \underset{\forall s \in N(t), \forall e \in E(s,t)}{\textbf{Aggregate}}\bigg(  \Big\{ \textbf{Extract}\big(H^{(l-1)}_s; H^{(l-1)}_t, e\big)\Big\} \bigg),
\end{align}
where $N(t)$ denotes all the source nodes of node $t$ and $E(s,t)$  denotes all the edges from node $s$ to $t$. 

There are two basic operators for GNNs, which are \textbf{Extract($\cdot$)} and \textbf{Aggregate($\cdot$)}. Among them,
\textbf{Extract($\cdot$)} represents the neighbor information extractor. It uses the target node's representation $H^{(l-1)}_t$ and the edge $e$ between the two nodes as query, and extract useful information from source node $H^{(l-1)}_s$. \textbf{Aggregate($\cdot$)} serves as the aggregation function of the neighborhood information. The \textit{mean, sum,} and \textit{max} functions are often considered as the basic aggregation operators, while sophisticated pooling and normalization functions can also be designed. Under this framework, various GNN architectures have been proposed.  
For example, the graph convolutional network (GCN) proposed by Kipf \textit{et al.}~\cite{gcn} averages the one-hop neighbor of each node in the graph, followed by a linear projection and then a non-linear activation. 
Hamilton \textit{et al.}~\cite{graphsage} propose GraphSAGE that generalizes GCN's aggregation operation from \textit{average} to \textit{sum, max} and a \textit{RNN unit}.

Also, there are a bunch of works incorporating the attention mechanism into GNNs. In general, the attention-based models implement the \textbf{Extract($\cdot$)} operation by estimating the importance of each source node, based on which a weighted aggregation is applied. For example, Velickovi \textit{et al.}~\cite{gat} propose the graph attention network (GAT), which adopts an additive mechanism to calculate attention and uses the same weight for calculating messages. 
Recently, Hu \textit{et al.} propose the heterogeneous graph transformer (HGT)~\cite{hgt} that leverages multi-head attentions for different relation types to get type-dependent attentions. 
The proposed pre-training framework \method\ can apply to all of these GNN models. 

\subsection{Pre-Training for Graphs} 

Previous studies have proposed to utilize pre-training to learn node representations, which largely belong to two categories. The first category is usually termed as network/graph embedding, which directly parameterizes the node embedding vectors and optimizes them by preserving some similarity measures, such as the network proximity~\cite{tang2015line} or statistics derived from random walks~\cite{grover2016node2vec,dong2017metapath2vec,qiu2018network}. 
However, the embeddings learned in this way cannot be used to initialize other models for fine-tuning over other tasks. 
In contrast, we consider a transfer learning setting, where the goal is to pre-train a generic GNN that can deal with different tasks. 

With the increasing focus on GNNs, researchers have explored the direction of pre-training GNNs on unannotated data. Kipf \textit{et al.} propose Variational Graph Auto-Encoders~\cite{DBLP:journals/corr/KipfW16a} to reconstruct the graph structure. Hamilton \textit{et al.} propose GraphSAGE~\cite{graphsage}, which can optimize via an unsupervised loss by using random walk based similarity metric. 
Velickovic \textit{et al.} introduce Graph Infomax~\cite{DBLP:journals/corr/abs-1809-10341}, which maximizes the mutual information between node representations obtained from GNNs and a pooled graph representation. 
Although these methods show enhancements over purely-supervised learning settings, the learning tasks can be  achieved by forcing nearby nodes to have similar embeddings, ignoring the rich semantics and higher-order structure of the graph. Our work proposes to pre-train GNNs by the permutated generative objective, which is a harder graph task and thus can guide the model to learn more complex semantics and structure of the input graph.

In addition, there are attempts to pre-train GNNs to extract graph-level representations. Sun \textit{et al.} present InfoGraph~\cite{infograph}, which maximizes the mutual information between graph-level representations obtained from GNNs and the representations of sub-structures. 
Hu \textit{et al.}~\cite{pretrain} introduce different strategies to pre-train GNNs at both node and graph levels and show that combining them together can improve the performance on graph classification tasks. 
Our work is different with them as our goal is to pre-train GNNs over a single (large-scale) graph and conduct the node-level transfer. 

\subsection{Pre-Training for Vision and Language} 


Pre-training has  been widely used in computer vision (CV) and natural language processing (NLP). 
In CV, early pre-training techniques~\cite{girshick2014rich,donahue2014decaf,DBLP:conf/cvpr/PathakKDDE16}  mostly follow the paradigm of first pre-training a model on large-scale supervised datasets (such as ImageNet~\cite{deng2009imagenet}) and then fine-tuning the pre-trained model on downstream tasks~\cite{girshick2014rich} or directly extracting the representations as features~\cite{donahue2014decaf}. Recently, some self-supervised tasks~\cite{DBLP:journals/corr/abs-1807-03748, he2019momentum, DBLP:journals/corr/abs-2002-05709} have also been utilized to pre-train vision models.
In NLP, Early works have been focused on learning (shallow) word embeddings~\cite{mikolov2013distributed,pennington2014glove} by leveraging the co-occurrence statistics on the text corpus. 
More recently, significant progresses have been made on contextualized word embeddings, such as BERT~\cite{DBLP:journals/corr/abs-1810-04805}, XLNET~\cite{xlnet} and GPT~\cite{radford2019language}. 
Take BERT as an example, it pre-trains a text encoder with two self-supervised tasks in order to better encode words and their contexts. 
These pre-training approaches have been shown to yield state-of-the-art performance in a wide range of NLP tasks and thus used as a fundamental component in many NLP systems.

\section{Generative Pre-Training of GNNs}\label{sec:approach}

In this section, we formalize the attributed graph generation task and introduce the generative pre-training framework (\method).

\subsection{The GNN Pre-Training Problem}

The input to GNNs is usually an attributed graph $G = (\cV, \cE, \cX)$, where $\cV$ and $\cE$ denote its node and edge sets, and $\cX$ represents the node feature matrix. 
A GNN model learns to output node representations under the supervision of a specific downstream task, such as node classification. Sometimes there exist multiple tasks on a single graph, and most GNNs require sufficient dedicated labeled data for each task. However, it is often challenging to obtain sufficient annotations, in particular for large-scale graphs, hindering the training of a well-generalized GNN. 
Therefore it is desirable to have a pre-trained GNN model that can generalize with few labels. 
Conceptually, this model should 1) capture the intrinsic structure and attribute patterns underlying the graph and 2) thus benefit various downstream tasks on this graph. 

\vpara{GNN Pre-Training.}Formally, our goal of GNN pre-training concerns the learning of a general GNN model $f_{\theta}$ purely based on single (large-scale) graph $G = (\cV, \cE, \cX)$ without labeled data such that $f_{\theta}$ is a good initialization for various (unseen) downstream tasks on the same graph or graphs of the same domain. 
To learn such a general GNN model without labeled data on the graph, a natural question arises here is: \textit{how to design an unsupervised learning task over the graph for pre-training the GNN model?}

\subsection{The Generative Pre-Training Framework}

\yd{if space allowed, to move the algo box from appendix here. So much text in this section, an algo flow will help a lot. Will come back after going through intro}

\hide{
Recent advances in self-supervised learning for NLP~\cite{DBLP:journals/corr/abs-1810-04805}, computer vision~\cite{DBLP:conf/cvpr/PathakKDDE16}, and speech~\cite{DBLP:journals/corr/abs-1807-03748}\kw{need a few referneces here}, we propose to design the pre-training task as self-supervised learning. 
Its promise is that we can learn an expressive model by predicting between parts of the input graph or its different views~\cite{}.

In this work, we propose to mask parts of the input graph (both its structure and node features) and predict/generate the masked structure and features based on the remaining part. 
To address this self-supervised learning task, we present the generative graph neural network pre-training model (\method). 
}

Recent advances in self-supervised learning for NLP~\cite{DBLP:journals/corr/abs-1810-04805,xlnet} and CV~\cite{DBLP:journals/corr/abs-1807-03748, he2019momentum, DBLP:journals/corr/abs-2002-05709} have shown that unlabeled data itself contains rich semantic knowledge, and thus a model that can capture the data distribution is able to transfer onto various downstream tasks. Inspired by this, we propose \method, which pre-trains a GNN by reconstructing/generating the input graph's structure and attributes. 

Formally, given an input graph $G = (\cV, \cE, \cX)$ and a GNN model $f_{\theta}$, we model the likelihood over this graph by this GNN as $p(G; \theta)$---representing how the nodes in $G$ are attributed and connected. \method\ aims to pre-train the GNN model by maximizing the graph likelihood, i.e., $\theta^* = \max_{\theta} p(G; \theta)$. 

Then, the first question becomes how to properly model $p(G; \theta)$. 
Note that most existing graph generation methods~\cite{DBLP:conf/icml/YouYRHL18, DBLP:conf/nips/LiaoLSWHDUZ19} follow the auto-regressive manner to factorize the probability objective, i.e., the nodes in the graph come in an order, and the edges are generated by connecting each new arriving node to existing nodes. 
Similarly, we denote a permutation vector $\pi$ to determine the node ordering, where $i^{\pi}$ denotes the node id of $i$-th position in permutation $\pi$. 
Consequently, the graph distribution $p(G; \theta)$ is equivalent to the expected likelihood over all possible permutations, i.e.,  
$$p(G; \theta) = \mathbb{E}_{\pi}\big[p_{\theta}(X^{\pi}, E^{\pi})\big],$$  
where $X^{\pi} \in \RR^{|\cV| \times d}$ denotes permutated node attributes and $E$ is a set of edges, while $E^{\pi}_i$ denotes  all edges connected with node $i^{\pi}$. For simplicity, we assume that observing any node ordering $\pi$ has an equal probability and also omit the subscript $\pi$ when illustrating the generative process for one permutation in the following sections. 
Given a permutated order, we can factorize the log likelihood autoregressively---generating one node per iteration---as:
\begin{align}\label{eq:gpt}
    \log p_{\theta}(X, E) = \sum_{i=1}^{|\cV|} \log p_{\theta}(X_{i}, E_{i} \mid X_{<i}, E_{<i}).
\end{align} 
\YS{(2) A small figure/table to illustrate these notations?}
At each step $i$, we use all nodes that are generated before $i$, their attributes $X_{<i}$, and 
the structure (edges) between these nodes $E_{<i}$ to generate a new node $i$, 
including both its attribute $X_{i}$ and its connections with existing nodes $E_{i}$. 

Essentially, the objective in Eq. \ref{eq:gpt} describes the autoregressive generative process of an attributed graph. 
The question becomes: \textit{how to model the conditional probability $p_{\theta}(X_{i}, E_{i} | X_{<i}, E_{<i})$?}

\hide{

By trying to predict the self-output from the self-input, you end up learning about the intrinsic properties / semantics of the object, which otherwise would have taken a ton of examples to learn from.

The principle is pretty simple: to encode an object, you try to setup learning tasks between parts of it or different views of it (the self).

generate output labels ‘intrinsically’ from data objects by exposing a relation between parts of the object, or different views of the object.
}

\begin{figure*}[ht!]
    \centering
    \includegraphics[width=1\textwidth]{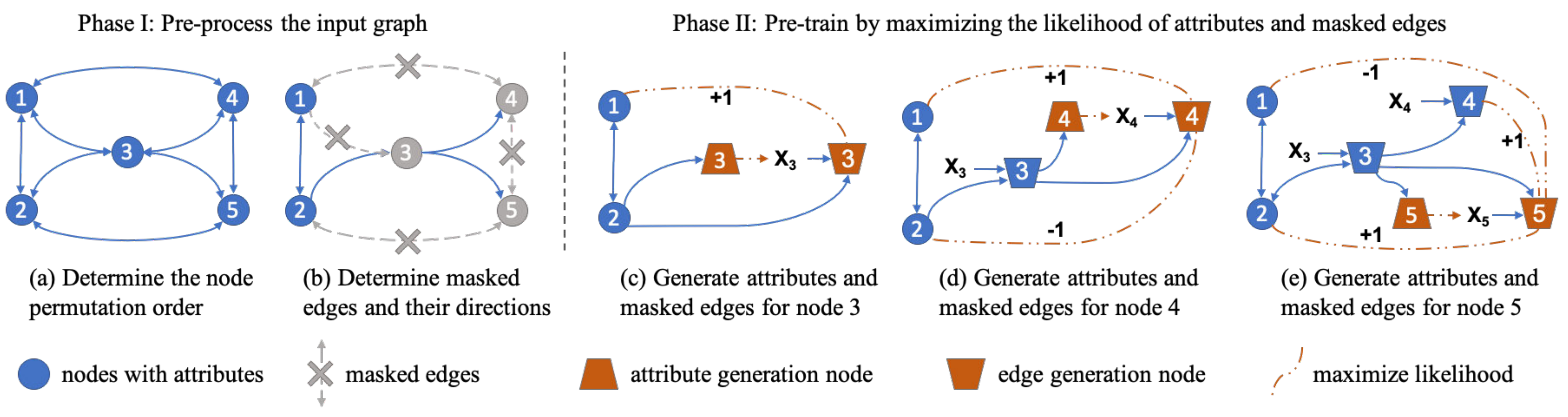}
    \caption{An illustrative example of the proposed attributed graph generation procedure.}
    \label{fig:framework}
\end{figure*}

\subsection{Factorizing Attributed Graph Generation}

To compute
$p_{\theta}(X_{i}, E_{i} | X_{<i}, E_{<i})$, one naive solution could be to simply assume that $X_{i}$ and $E_{i}$ are independent, 
that is, 
$$p_{\theta}(X_{i}, E_{i} | X_{<i}, E_{<i}) = p_{\theta}(X_{i} | X_{<i}, E_{<i}) \cdot p_{\theta}(E_{i} | X_{<i}, E_{<i})$$
With such decomposition, for each node, the dependency between its attributes and connections are completely neglected. 
However, the ignored dependency is the core property of attributed graphs and also the foundation of convolutional aggregation in GNNs. Therefore, such a naive decomposition cannot provide informative guidance for pre-training GNNs. 

To address this issue, we present the dependency-aware factorization mechanism for the attributed graph generation process.  
Specifically, when estimating a new node's attributes, we are given its structure information, and vice versa. 
During the process, a part of the edges has already been observed (or generated). 
Then the generation can be decomposed into two coupled parts: 
\begin{itemize}
\item given the observed edges, generate node attributes; 
\item given the observed edges and generated node attributes, generate the remaining edges. 
\end{itemize}
In this way, the model can capture the dependency between the attributes and structure for each node. 

Formally, we define a variable $o$ to denote the index vector of all the observed edges within $E_i$. Thus, $E_{i, o}$ denotes the observed edges. 
Similarly, $\neg o$ denotes the index of all the masked edges, which are to be generated. 
With this, we can rewrite the conditional probability as an expected likelihood over all observed edges: 
\begin{align}
   &  p_{\theta}(X_{i}, E_{i} \mid X_{<i}, E_{<i}) \nonumber\\
= &\sum\nolimits_{o} p_{\theta}(X_{i}, E_{i, \neg o} \mid E_{i, o}, X_{<i}, E_{<i}) \cdot p_{\theta}(E_{i, o} \mid X_{<i}, E_{<i}) \nonumber\\
= &\mathbb{E}_{o}\Big[p_{\theta}(X_{i}, E_{i, \neg o} \mid E_{i, o}, X_{<i}, E_{<i})\Big] \nonumber \\ 
=&\mathbb{E}_{o}\Big[\underbrace{
 p_{\theta}(X_{i} \mid E_{i, o}, X_{<i}, E_{<i})
 }_\text{1) generate attributes} \cdot 
 \underbrace{
 p_{\theta}(E_{i, \neg o} \mid E_{i, o}, X_{\leq i}, E_{<i})
 }_\text{2) generate edges}\Big]. \label{eq:overall}
\end{align}

\YS{The above factorization is not exact. We also need to multiply $p(A_{i,<t}|...)$. In order to get rid of this item, i.e., how to generate the first t links, we need to use sampling to approximate it, which has to be mentioned. But how to sample first t links given a permutation? It should already fixed. So I still have question on how to compute $p(A_{i,<t}|...)$. Of course, we can always downgrade our claim saying we are optimizing the conditional distribution of X and remaining edges given observed edges. }
\zn{Determine $\tau$ and $t$ itself is a biased sample. We want the sampling to reflect the probability of whether a node should be observed link. I write it in the expectation (already rewrite it). We can easily implement it by sorting it by degree or treat it as uniform.}
\kw{This notation is better, but I still feel we should prevent presenting the framework in this way as it is too general. We are not actually modeling $p_{\theta}(E_{i, o} \mid X_{<i}, E_{<i})$ and just approximate it by sampling. Decomposing the probability in this way, we hide the essential component to the expectation symbol and the readers will get confused. My main concern is that generating a whole graph from scratch sounds unrealistic and we do not have a good answer if it is possible. It may confuse the reviewers that you're trying to use some unrealistic objective to train a model. In contrast, the setting in your experiments makes much more sense.}\YS{For the second line, what is the meaning of sum over all the possible observations? How we come from first line to second line?}\zn{Since o is a intermediate variable, to marginalize the probability, we need to sum over all possible observation with its probability, following \url{https://en.wikipedia.org/wiki/Law_of_total_expectation}}

This factorization design is able to model the dependency between node $i$'s attributes $X_{i}$ and its associated connections $E_{i}$. 
The first term $p_{\theta}(X_{i} \mid E_{i, o}, X_{<i}, E_{<i})$ denotes the generation of attributes for node $i$. 
Based on the observed edges $E_{i, o}$, we gather the target node $i$'s neighborhood information to generate its attributes $X_{i}$. 
The second term $p_{\theta}(E_{i, \neg o} \mid E_{i, o}, X_{\leq i}, E_{<i})$ denotes the generation of masked edges. 
Based on both the observed edges $E_{i, o}$ and the generated attributes $X_{i}$, we generate the representation of the target node $i$ and predict whether each edge within $E_{i, \neg o}$ exists. 

\vpara{A graph generation example.}
We intuitively show how the proposed factorization-based graph generation process works. Take, for example, an academic graph, if we would like to generate one paper node, whose title is considered as its attribute, while this paper node is connected to its authors, published venue, and cited papers. 
Based on some observed edges between this paper and some of its authors, our generation process first generates its title. Then,
based on both the observed edges and generated title, we predict its remaining authors, published venue, and references. 
In this way, this process models the interaction between the paper's attribute (title) and structure (observed and remaining edges) to complete the generation task, bringing in informative signals for pre-training GNNs over the academic graph.

So far, we factorize the attributed graph generation process into a node attribute generation step and an edge generation step. 
The question we need to answer here is: \textit{How to efficiently pre-train GNNs by optimizing both attribute and edge generation tasks?}

\subsection{Efficient Attribute and Edge Generation}
\label{subsec:attr-edge-nodes}





For the sake of efficiency, it is desired to compute the loss of attribute and edge generations by running the GNN only once for the input graph. In addition, we expect to conduct attribute generation and edge generation simultaneously. However, edge generation requires node attributes as input, which can be leaked to attribute generation. To avoid information leakage, we design to separate each node into two types:
\begin{itemize}
    \item Attribute Generation Nodes. We mask out the attributes of these nodes by replacing their attributes with a dummy token and learn a shared vector $X^{init}$ to represent it\footnote{$X^{init}$ has the same dimension as $X_{i}$ and can be learned during pre-training.}. This is equivalent to the trick of using the [Mask] token in the masked language model~\cite{DBLP:journals/corr/abs-1810-04805}.

    \item Edge Generation Nodes. For these nodes, we keep their attributes and put them as input to the GNN.

\end{itemize}
We then input the modified graph to the GNN model and generate the output representations. We use $h^{Attr}$ and $h^{Edge}$  to represent the output embeddings of Attribute Generation and Edge Generation Nodes, respectively. As the attributes of Attribute Generation Nodes are masked out,  $h^{Attr}$ in general contains less information than  $h^{Edge}$. Therefore, when conduct the GNN message passing, we only use Edge Generation Nodes' output $h^{Edge}$ as outward messages. The representations of the two sets of nodes are then used to generate attributes and edges with different decoders.


For Attribute Generation, we denote its decoder as ${Dec}^{Attr}(\cdot)$, which takes $h^{Attr}$ as input and generates the masked attributes. The modeling choice depends on the type of attributes. For example, if the input attribute of a node is text, we can use the text generator model (e.g., LSTM) to generate it. If the input attribute is a standard vector, we can apply a multi-layer Perceptron to generate it. Also, we define a distance function as a metric between the generated attributes and the real ones, such as perplexity for text or L2-distance for vectors. Thus, we calculate the attribute generation loss via:
\begin{align}
\mathcal{L}^{Attr}_i = Distance\big({Dec}^{Attr}(h^{Attr}_i), X_{i}\big). \label{eq:gen}
\end{align}

By minimizing the distance between the generated and masked attributes, it is equivalent to maximize the likelihood to observe each node attribute, i.e., $ p_{\theta}(X_{i} \mid E_{i, o}, X_{<i}, E_{<i})$, and thus the pre-trained model can capture the semantic of this graph.

For Edge Generation, we assume that the generation of each edge is independent with others, so that we can factorize the likelihood:
\begin{align}
    p_{\theta}(E_{i, \neg o} \mid E_{i, o}, X_{\leq i}, E_{<i}) = \prod_{j^{+} \in E_{i, \neg o}} \! p_{\theta}(j^{+} \mid E_{i, o}, X_{\leq i}, E_{<i}). \label{eq:obj}
\end{align}

Next, after getting the Edge Generation node representation $h^{Edge}$, we model the likelihood that node $i$ is connected with node ${j}$ by $Dec^{Edge}(h^{Edge}_i, h^{Edge}_{j})$, where $Dec^{Edge}$ is a pairwise score function. Finally, we adopt the negative contrastive estimation to calculate the likelihood for each linked node $j^+$. We prepare all the unconnected nodes as $S^{-}_i$ and calculate the contrastive loss via
\begin{align}
\label{eq:contrastive}
\mathcal{L}^{Edge}_i \!= -\!\!\sum_{j^{+} \in E_{i, \neg o}}\log \frac{\exp\big(Dec^{Edge}(h^{Edge}_i,\  h^{Edge}_{j^{+}})\big)}{\sum_{j^{} \in S^{-}_i \cup \{ j^{+}\} }\exp\big(Dec^{Edge}(h^{Edge}_i,\ h^{Edge}_{j^{}})\big)}
\end{align}

By optimizing $\mathcal{L}^{Edge}$, it is equivalent to maximizing the likelihood of generating all the edges, and thus the pre-trained model is able to capture the intrinsic structure of the graph.

\YS{What I expected is to see how to compute $p(remaining edges|observed edges, features)$. What is provided here is a loss. How to connect this loss to probability? As a generative model, we have to provide probabilities.}\zn{(1) I change the previous notation from A -> E, denoting only linked edges. In this way, the problem becomes a standard node embedding learning problem, and the NCE Loss we use can well capture the probability (2) I've changed the writing for these two tasks, does it make sense now?}

Figure~\ref{fig:framework} illustrates the attributed graph generation process. Specifically:
(a) We determine the node permutation order $\pi$ for the input graph. (b) We randomly select a portion of the target node's edges as observed edges $E_{i, o}$ and the remaining as masked edges $E_{i, \neg o}$ (grey dashed lines with cross). We delete masked edges in the graph.
(c) We separate each node into the Attribute Generation and Edge Generation nodes to avoid information leakage.
(d) After the pre-processing, we use the modified adjacency matrix to calculate the representations of node 3,4 and 5, including both their Attribute and Edge Generation Nodes. 
Finally, as illustrated in (d)--(e), we train the GNN model via the attribute prediction and masked edge prediction task for each node in parallel. The overall pipeline of \method\ is illustrated in Algo.~\ref{alg:pipeline} (See Appendix B for details).

\begin{algorithm}[ht!] 
\caption{The \method\ Pre-Training Framework} 
\label{alg:pipeline} 
\begin{algorithmic}[1] 
\REQUIRE
Input Attributed Graph $G$, Graph Sampler $Sampler(\cdot)$.\\
\ENSURE
\STATE Initialize the GNN model as $f_{\theta}$, the attribute generation decoder as ${Dec}^{Attr}$, and the edge generation decoder as ${Dec}^{Edge}$.
\STATE Initialize the adaptive node embedding queue $Q=\{\}$ and the attribute vector $h^{init}$.
 \FOR {each sampled graph $\hat{G} \in Sampler(G)$}
    \STATE For each node, sample the observed edge index $o$ and masked edges $\neg o$, and delete masked edges $E_{i, \neg o}$ accordingly. \label{alg:step2}
    \STATE Separate each node into the Attribute Generation and Edge Generation nodes. Replace the input to Attribute Generation node as $h_{init}$. Apply GNN $f_{\theta}$ to get two sets of node embeddings $h^{Attr}$ and $h^{Edge}$ for each node in the graph. \label{alg:step3}
    \FOR {node $i$ with attributes $X_{i}$ and masked edges $E_{i, \neg o}$}
        \STATE Calculate the attribute generation loss $\mathcal{L}^{Attr}$ by Eq. \ref{eq:gen} \label{alg:step4}
        \STATE Prepare negative samples $S^{-}_i$ for edge generation by concatenating unconnected nodes and adaptive queue $Q$.
        \label{alg:step9}\STATE Calculate the edge generation loss $\mathcal{L}^{Edge}$ by Eq. \ref{eq:contrastive} \label{alg:step5}
    \ENDFOR
    \STATE Optimize $\theta$ by minimizing $\mathcal{L}^{Attr}$ and $\mathcal{L}^{Edge}$. \label{alg:step6}
    \STATE Update $Q$ by adding in $h^{Edge}$ and popping out most outdated embeddings. \label{alg:step7}
\ENDFOR
\RETURN Pre-trained model parameters $\theta^*$ for downstream tasks
\end{algorithmic} 
\end{algorithm}

\subsection{\hspace{-0.3cm}\method\ for Heterogeneous \& Large  Graphs}

In this section, we discuss how to apply \method\ to pre-train for large-scale and heterogeneous graphs, which can be of practical use for modeling real-world complex systems~\cite{Sun:BOOK2012,dong2020HeterNRL}, such as academic graphs, product graphs, IoT networks, and knowledge graphs.

\vpara{Heterogeneous graphs.}
Many real-world graphs are heterogeneous, meaning that they contain different types of nodes and edges. 
For heterogeneous graphs, the proposed \method\ framework can be straightforwardly applied to pre-train heterogeneous GNNs. 
The only difference is that each type of nodes and edges may have its own decoder, which is specified by the heterogeneous GNNs rather than  the pre-training framework. 
All the other components remain exactly the same.

\vpara{Large-scale graphs.}
To pre-train GNNs on graphs that are too large to fit into the hardware, we sample subgraphs for training. 
In particular, we propose to sample a dense subgraph from homogeneous and heterogeneous graphs by using the LADIES algorithm~\cite{ladies} and its heterogeneous version HGSampling~\cite{hgt}, respectively. 
Both methods theoretically guarantee that the sampled nodes are highly interconnected with each other and maximally preserve the structural information.

To estimate the contrastive loss in Eq. \ref{eq:contrastive}, it is required to go over all nodes of the input graph. 
However, we only have access to the sampled nodes in a subgraph for estimating this loss, making the (self-)supervision only focus on local signals.
\yd{to kw and zn, plz check whether this would work}
\kw{The motivation here is a bit not clear. You may want to elaborate more on why the task becomes easier and why the guidance is weaker and why the adaptive queue can resolve this issue.} 
To alleviate this issue, we propose the \textit{Adaptive Queue}, which stores node representations in previously-sampled subgraphs as negative samples. 
Each time we process a new subgraph, we progressively update this queue by adding the latest node representations and remove the oldest ones. As the model parameters will not be updated rigorously, the negative samples stored in the queue are consistent and accurate. 
The \textit{Adaptive Queue} enables us to use much larger negative sample pools $S^{-}_i$. Moreover, the nodes across different sampled sub-graphs can bring in the global structural guidance for contrastive learning.


\section{Evaluation}\label{sec:evaluation}

To evaluate the performance of \method, we conduct experiments on the Open Academic Graph (OAG) and Amazon Recommendation datasets.
To evaluate the generalizability of \method, we consider different transfer settings---time transfer and field transfer---which are of practical importance. 


\subsection{Experimental Setup}

\vpara{Datasets and Tasks.}We conduct experiments on both heterogeneous and homogeneous graphs. 
For heterogeneous graphs, we use 
the Open Academic Graph and Amazon Review Recommendation data. 
For homogeneous graphs, we use the Reddit dataset~\cite{graphsage} and the paper citation network extracted from OAG. 
All datasets are publicly available and the details can be found in Appendix A.

\textit{Open Academic Graph (OAG)}~\cite{wang2020mag,DBLP:conf/kdd/ZhangLTDYZGWSLW19,tang2008arnetminer} contains more than 178 million nodes and 2.236 billion edges. It is the largest publicly available heterogeneous academic dataset to date. Each paper is labeled with a set of research topics/fields (e.g., Physics and Medicine) and the publication date ranges from 1900 to 2019. 
We consider the prediction of Paper--Field, Paper--Venue, and Author Name Disambiguation (Author ND) as three downstream tasks~\cite{hgt,dong2020HeterNRL}. The performance is evaluated by MRR---a widely adopted ranking metric~\cite{DBLP:books/daglib/0027504}.

\textit{Amazon Review Recommendation Dataset (Amazon)}~\cite{DBLP:conf/emnlp/NiLM19}  contains 82.8 million reviews, 20.9 million users, and 9.3 million products. 
The reviews are published from 1996 to 2018. 
Each review consists of a discrete rating score from 1 to 5 and a specific field, including book, fashion, etc. For downstream tasks, we predict the rating score as a five-class classification task within the Fashion, Beauty, and Luxury fields. We use micro F1-score as the evaluation metric.



\vpara{The base GNN model.} On the OAG and Amazon datasets, we use the state-of-the-art heterogeneous GNN---Heterogeneous Graph Transformer (HGT)~\cite{hgt}---as the base model for \method. 
Furthermore, we also use other (heterogeneous) GNNs as the base model to test our generative pre-training framework. 


\vpara{Implementation details.}For all base models, we set the hidden dimension 
as 400, the head number as 8, and the number of GNN layers as 3. 
All of them are implemented using the PyTorch Geometric (PyG) package~\cite{pyG}.

We optimize the model via the AdamW optimizer~\cite{DBLP:conf/iclr/LoshchilovH19} with the Cosine Annealing Learning Rate Scheduler~\cite{DBLP:conf/iclr/LoshchilovH17} with 500 epochs and select the one with the lowest validation loss as the pre-trained model. 
We set the adaptive queue size to be 256. 

During downstream evaluation, we fine-tune the model using the same optimization setting for 200 epochs as that in pre-training. We train the model on the downstream tasks for five times and report the mean and standard deviation of test performance.

\vpara{Pre-training baselines.}There exist several works that propose unsupervised objectives over graphs, which can potentially be used to pre-train GNNs. We thus compare the proposed \method\ framework with these baselines:\begin{itemize}
    \item \textit{GAE}~\cite{DBLP:journals/corr/KipfW16a}, which denotes graph auto-encoders, focuses on a traditional link prediction task. It randomly masks out a fixed proportion of the edges and asks the model to reconstruct these masked edges.
    \item \textit{GraphSAGE (unsp.)}~\cite{graphsage} forces connected nodes to have similar output node embeddings. Its main difference with GAE lies in that it does not mask out the edges during  pre-training.
    \item \textit{Graph Infomax}~\cite{DBLP:journals/corr/abs-1809-10341} tries to maximize the local node embeddings with global graph summary embeddings. Following its setting for a large-scale graph, for each sampled subgraph, we shuffle the graph to construct negative samples. 
\end{itemize}
In addition, we also evaluate the two pre-training tasks in \method by using each one of them alone, that is, attribute generation---\textit{\method (Attr)}---and edge generation---\textit{\method (Edge)}.



\begin{table*}[!th]
\centering
\begin{tabular}{llcccccccccc}
\toprule
 & {Downstream Dataset} &&  \multicolumn{3}{c}{OAG} & & & \multicolumn{3}{c}{Amazon}  & \\ \cmidrule(lr){1-2}  \cmidrule(lr){4-6} \cmidrule(lr){9-11}
& {Evaluation Task}    && Paper--Field    & Paper--Venue    & Author ND & & & Fashion & Beauty & Luxury \\ \toprule
~ & No Pre-train && .336$\pm$.149 & .365$\pm$.122    &  .794$\pm$.105 &&& .586$\pm$.074 & .546$\pm$.071 & .494$\pm$.067 \\ \toprule
\multirow{6}{*}{\rotatebox[origin=c]{90}{{Field Transfer}}}  & GAE && .403$\pm$.114  &  .418$\pm$.093   &  .816$\pm$.084  &&&  .610$\pm$.070  & .568$\pm$.066 & .516$\pm$.071     \\ 
~ & GraphSAGE (unsp.)&& .368$\pm$.125  &  .401$\pm$.096   &  .803$\pm$.092 &&&  .597$\pm$.065  & .554$\pm$.061 & .509$\pm$.052      \\ 
~ & Graph Infomax && .387$\pm$.112  &  .404$\pm$.097   &  .810$\pm$.084  &&& .604$\pm$.063  & .561$\pm$.063 & .506$\pm$.074    \\ \cmidrule{2-11}
~ & \method\ (Attr) & &   .396$\pm$.118   &  .423$\pm$.105 &  .818$\pm$.086   && &  .621$\pm$.053  &   .576$\pm$.056   &  .528$\pm$.061      \\
~ & \method\ (Edge) &&   .401$\pm$.109  &   .428$\pm$.096  &  .826$\pm$.093 && &   .616$\pm$.060  &   .570$\pm$.059  & .520$\pm$.047             \\
~ & \method\   &&   \textbf{.407$\pm$.107}    &  \textbf{.432$\pm$.098}    &   \textbf{.831$\pm$.102}    &  & &\textbf{.625$\pm$.055}   &   \textbf{.577$\pm$.054}    &  \textbf{.531$\pm$.043} \\ \toprule
\multirow{6}{*}{\rotatebox[origin=c]{90}{{Time Transfer}}}  & GAE && .384$\pm$.117  &  .412$\pm$.101   &  .812$\pm$.095 & &&  .603$\pm$.065  & .562$\pm$.063 & .510$\pm$.071     \\ 
~ & GraphSAGE (unsp.)&& .352$\pm$.121  &  .394$\pm$.105   &  .799$\pm$.093 &&&  .594$\pm$.067  & .553$\pm$.069 & .501$\pm$.064      \\ 
~ & Graph Infomax && .369$\pm$.116  &  .398$\pm$.102   &  .805$\pm$.089  & && .599$\pm$.063  & .558$\pm$.060 & .503$\pm$.063    \\ \cmidrule{2-11}
~ & \method\ (Attr) & &   .382$\pm$.114   &  .414$\pm$.098 &  .811$\pm$.089   && &  .614$\pm$.057  &   \textbf{.573$\pm$.053}   &  .522$\pm$.051      \\
~ & \method\ (Edge) &&   .392$\pm$.105  &   .421$\pm$.102  &  .821$\pm$.088 && &   .608$\pm$.055  &   .567$\pm$.038  & .513$\pm$.058             \\
~ & \method\   &&   \textbf{.400$\pm$.108}    &  \textbf{.429$\pm$.101}    &   \textbf{.825$\pm$.093}    &  & &\textbf{.617$\pm$.059}   &   .572$\pm$.059    &  \textbf{.525$\pm$.057} \\ \toprule

\multirow{8}{*}{\rotatebox[origin=c]{90}{{Time + Field Transfer}}} & GAE && .371$\pm$.124  &  .403$\pm$.108   &  .806$\pm$.102 & &&  .596$\pm$.065  & .554$\pm$.063 & .505$\pm$.061     \\ 
~ & GraphSAGE (unsp.) && .349$\pm$.130  &  .393$\pm$.118   &  .797$\pm$.097&  &&  .589$\pm$.071  & .545$\pm$.068 & .498$\pm$.064      \\ 
~ & Graph Infomax && .360$\pm$.121  &  .391$\pm$.102   &  .800$\pm$.093  & && .591$\pm$.068  & .550$\pm$.058 & .501$\pm$.063    \\ \cmidrule{2-11}
~ & \method\ (Attr) & &   .364$\pm$.115   &  .409$\pm$.103 &  .809$\pm$.094   && &  .608$\pm$.062  &   .569$\pm$.057   &  .517$\pm$.057      \\
~ & --- (w/o node separation) & &   .347$\pm$.128   &  .391$\pm$.102 &  .791$\pm$.108   && &  .585$\pm$.068  &   .546$\pm$.062   &  .497$\pm$.062      \\
~ & \method\ (Edge) &&   .386$\pm$.116  &   .414$\pm$.104  &  .815$\pm$.105 && &   .604$\pm$.058  &   .565$\pm$.057  & .514$\pm$.047             \\
~ & --- (w/o adaptive queue) &&   .376$\pm$.121  &   .410$\pm$.115  &  .808$\pm$.104 && &   .599$\pm$.068  &   .562$\pm$.065  & .509$\pm$.062             \\
~ & \method\   &&   \textbf{.393$\pm$.112}    &  \textbf{.420$\pm$.108}    &   \textbf{.818$\pm$.102}    &  & &\textbf{.610$\pm$.054}   &   \textbf{.572$\pm$.063}    &  \textbf{.521$\pm$.049} \\ \bottomrule
\end{tabular} 
\caption{Performance of different downstream tasks on OAG and Amazon by using different pre-training frameworks with the heterogeneous graph transformer (HGT)~\cite{hgt} as the base model. 
10\% of labeled data is used for fine-tuning. 
} 
\label{tab:result}
\end{table*}

\subsection{Pre-Training and Fine-Tuning Setup}

The goal of pre-training is to transfer knowledge learned from numerous unlabeled nodes of a large graph to facilitate the downstream tasks with a few labels. Specifically, we first pre-train a GNN and use the pre-trained model weights to initialize models for downstream tasks. We then fine-tune the models with the downstream task specific decoder on the training (fine-tuning) set and evaluate the performance on the test set. 

Broadly, there are two different setups. 
The first one is to pre-train and fine-tune on exactly the same graph. 
The second one is relatively more practical, which is to pre-train on one graph and fine-tune on unseen graphs of the same type as the pre-training one. 
Specifically, we consider the following three graph transfer settings between the pre-training and fine-tuning stages: 
\begin{itemize}
    \item \textit{Time Transfer}, where we use data from different time spans for pre-training and fine-tuning.
    For both OAG and Amazon, we use data before 2014 for pre-training and data since 2014 for fine-tuning. 
    
    \item \textit{Field Transfer}, where we use data from different fields for pre-training and evaluating. 
    In OAG, we use papers in the field of computer science (CS) for downstream fine-tuning and use all  papers in the remaining fields (e.g., Medicine) for pre-training. 
    In Amazon, we pre-train on products in Arts, Crafts, and Sewing, and fine-tune on products in Fashion, Beauty, and Luxury. 
    
    \item \textit{Time + Field Transfer}, where we use the graph of particular fields before 2014 to pre-train the model and use the data from other fields since 2014 for fine-tuning. 
    Intuitively, this combined transfer setting is more challenging than the transfer of time or field alone. 
    For example, we pre-train on the OAG graph except CS field before 2014 and fine-tune on the CS graph since 2014.  
\end{itemize}

During fine-tuning, for both datasets, we choose nodes from 2014 to 2016 for training, 2017 for validation, and since 2018 for testing. 
To meet the assumption that training data is usually scarce, we only provide 10\% of the labels for training (fine-tuning) by default, while the ablation study over different data percentages is also conducted.
During pre-training, we randomly select a subset of the data as the validation set. 



\subsection{Experimental Results}

We summarize the performance of downstream tasks with different pre-training methods on OAG and Amazon in Table ~\ref{tab:result}. 
As discussed above, we setup three different transfer settings between pre-training and fine-tuning stages: Field Transfer, Time Transfer, and Field + Time Combined Transfer, as organized in three different blocks in the Table. 

Overall, the proposed \method\ framework significantly enhances the performance for all downstream tasks on both datasets. 
On average, \method achieves relative performance gains of  13.3\% and 5.7\% over the base model without pre-training on OAG and Amazon, respectively. 
Moreover, it consistently outperforms other pre-training frameworks, such as Graph Infomax, across different downstream tasks for all three transfer settings on both datasets.


\vpara{Different transfer settings.}Observed from Table \ref{tab:result}, the performance gain lifted by pre-training under the field transfer is higher than that under the time transfer, and the time + field combined transfer is the most challenging setting as evident in the least performance gain brought by pre-training. 
Nonetheless, under the combined transfer, \method\ still achieves 11.7\% and 4.6\% performance gains on both datasets, respectively. 
Altogether, the results suggest that \textit{the proposed generative pre-training strategy enables the GNN model to capture the generic structural and semantic knowledge of the input graph, which can be used to fine-tune on the unseen part of the graph data. }

\vpara{Ablation studies on pre-training tasks.}We analyze the effectiveness of the two pre-training tasks in \method---attribute generation and edge generation---by examining which of them is more beneficial for the pre-training framework and, by extension, downstream tasks. 
In Table \ref{tab:result}, we report the performance of \method by using attribute generation and edge generation alone, that is, \method (Attr) and \method (Edge). 
On OAG, the average performance gains by \method (Attr) and \method (Edge) are 7.4\% and 10.3\%, respectively, suggesting that Edge Generation is a more informative pre-training task than Attribute Generation in \method. 
However, we have an opposite observation for Amazon, on which the performance improved by Attribute Generation is 5.2\% in contrast to the 4.1\% improvement lifted by Edge Generation.  
\textit{This suggests that the \method framework benefits differently from attribute and edge generations on different datasets. 
However, combining the two pre-training tasks together produces the best performance on both cases. 
}

We further compare the Edge Generation task against other edge-based pre-training methods---GAE and GraphSage (unsp.)---in Table \ref{tab:result}. 
On OAG, the performance improvements brought by \method's edge generation, GAE, and GraphSage over no pre-training are 10.3\%, 7.4\%, and 4.0\%, respectively. 
On Amazon, the gains are 5.2\%, 3.1\%, and 1.3\%, respectively. 
From the comparisons, we have the following observations. 
First, both GAE and \method's edge generation offer better results than GraphSage on both datasets, demonstrating that 
masking on edges is an effective strategy for self-supervised graph representation learning. 
Without edge masking, the model simply retains a similar embedding for connected nodes, as the label we would like to predict (whether two nodes are linked) has already been encoded in the input graph structure. Such information leakage will downgrade the edge prediction task to a trivial problem. 
Second, the proposed Edge Generation task consistently outperforms GAE. 
The main advantage of \method's edge generation comes from that 
it learns to generate missing edges autoregressively and thus can capture the dependencies between the masked edges, which are discarded by GAE. 
\textit{In summary, the results suggest that the proposed graph generation tasks can give informative self-supervision for GNN pre-training.}

\begin{table}[!t]
\centering
\begin{tabular}{lccccc}
\toprule
Model  & HGT & GCN & GAT & RGCN  & HAN \\ \midrule
No Pre-train & \textbf{.336} & .317   &  .308  &    .296 &    .322\\ 
\method  & \textbf{.407} & .349   &  .362  &    .351 &    .384\\ 
Relative Gain &   21.1\% & 10.1\%   &  17.5\%  &    18.6\% &    19.3\%\\ 
\bottomrule
\end{tabular} 
\caption{Compare the pre-training Gain with different GNN architectures. Evaluate on OAG, Paper-Field Task, under Combined Transfer setting with 10\% training data.} 
\label{tab:gnn}
\vspace{-0.25cm}
\end{table}

\vpara{Ablation studies on the base GNN.}
We investigate whether the other GNN architectures can benefit from the proposed pre-training framework. 
Therefore, in addition to HGT~\cite{hgt}, we try
GCN~\cite{gcn}, GAT~\cite{gat}, RGCN~\cite{DBLP:conf/esws/SchlichtkrullKB18}, and HAN~\cite{DBLP:conf/www/WangJSWYCY19} as the base model. 
Specifically, we pre-train them on OAG and then use the paper-field prediction task under the combined transfer setting with 10\% of training data for fine-tuning and testing. 
Model-independent hyper-parameters, such as the hidden dimension size and optimization, are set the same. 
The results are reported in in Table \ref{tab:gnn}. 
We can observe that 1) HGT achieves the best performance among all non pre-trained GNN models; 
2) \method with HGT generates the most promising results for the concerned downstream task; 
and 3) \textit{more importantly, the proposed \method pre-training framework can enhance the downstream performance for all the GNN architectures.}

\vpara{Ablation studies on the node separation and adaptive queue.} Finally, we examine the effectiveness of the two design choices of \method, i.e., node separation and adaptive queue. 
The node separation is designed for alleviating the information leakage problem for the Attribute Generation task. 
Without this component, the attributes would appear in the input and thus the pre-training method would only need to maintain the input features for output. 
In other words, it cannot learn any knowledge of the input graph that could be transferred to downstream tasks and thus affect the results negatively. 
From Table \ref{tab:result}, we can see that 
the attribute generation based pre-training model suffers from the removal of the node separation component (w/o node separation), 
and in many cases, its performance is even worse than the ones without pre-training. 
\textit{This demonstrates the significance of this node separation design in avoiding
attribute information leakage. }

The adaptive queue is designed for alleviating the gap between the sampled graphs and the full graph. 
Similarly, we conduct the ablation study by removing it from the Edge Generation based pre-training model and from Table \ref{tab:result}, we witness the consistent performance drops for all tasks---\method (Edge) vs. (w/o adaptive queue). 
\textit{This indicates that adding more negative samples by using the adaptive queue is indeed helpful to the pre-training framework.}

\begin{figure}[t!]
    \centering
    \includegraphics[width=0.475\textwidth, trim = 0 0 10 10, clip]{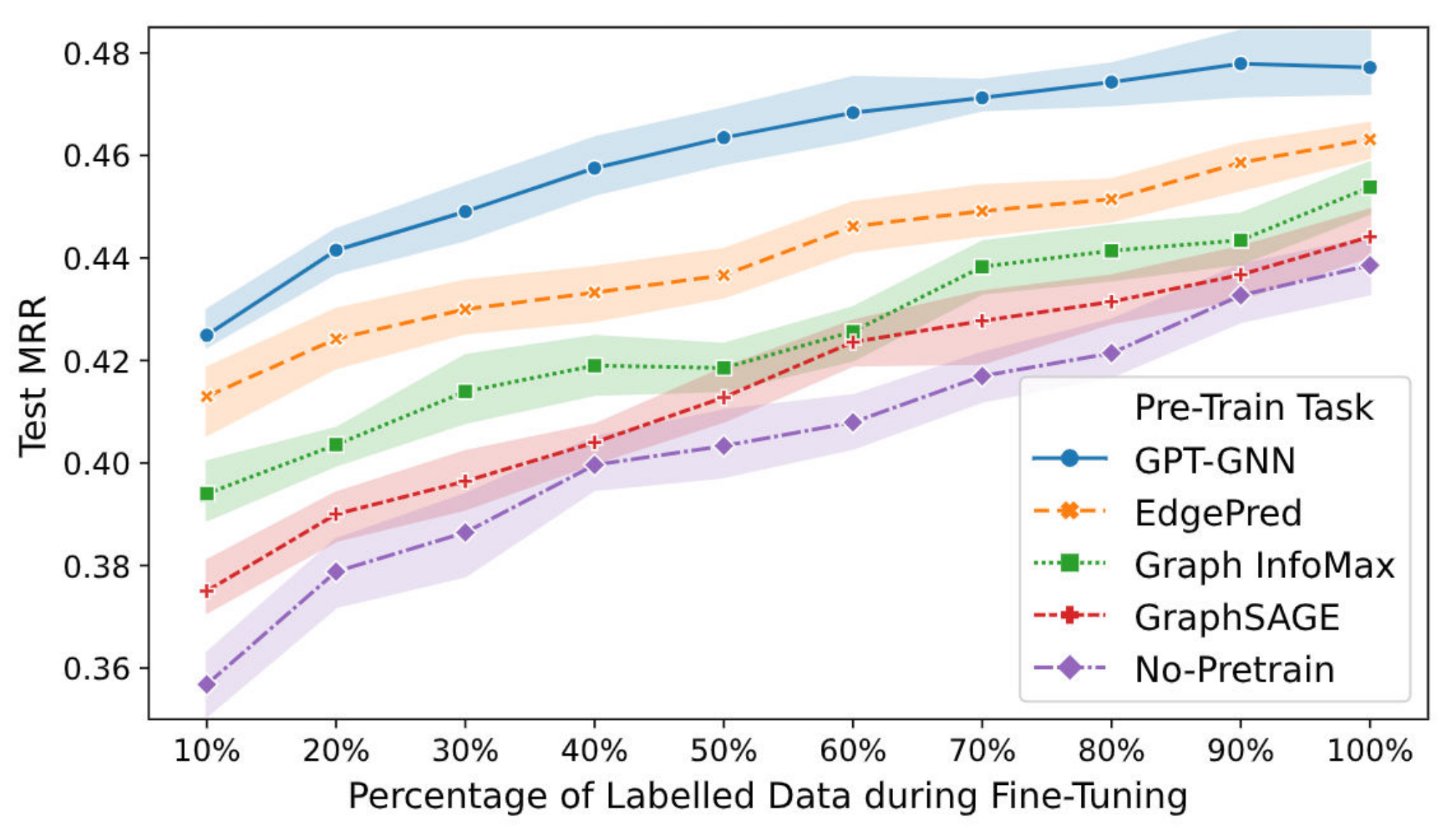}
    \vspace{-0.65cm}
    \caption{Compare pre-training tasks with different training data size. 
    Evaluated by the paper--field prediction task on OAG under the field transfer setting.}
    \label{fig:percentage}
    \vspace{0.3cm}
\end{figure}

\begin{table}[!t]
\centering
\begin{tabular}{lcc}
\toprule
Downstream Dataset  & OAG (citation) & Reddit \\ \midrule
No Pre-train & .281$\pm$.087 & .873$\pm$.036   \\  \midrule
GAE & .296$\pm$.095 &.885$\pm$.039   \\ 
GraphSAGE (unsp.)  & .287$\pm$.093 & .880$\pm$.042  \\ 
Graph Infomax  & .291$\pm$.086 & .877$\pm$.034   \\ 
\midrule
\method  & \textbf{.309$\pm$.081} & \textbf{.896$\pm$.028}  \\ 
\bottomrule
\end{tabular} 
\caption{Downstream performance on homogeneous graphs, including the paper citation network in OAG and Reddit. } 
\label{tab:homo}
\vspace{-0.25cm}
\end{table}

\vpara{Training data size.}In Figure~\ref{fig:percentage}, we examine whether the proposed \method\ method can generalize well with different training data size during fine-tuning, i.e., from 10\% to 100\%. 
First, we can observe that \method\ and other pre-training frameworks consistently improve the downstream task performance with more labeled training data. 
Second, it is clear that \method performs the best among all pre-training tasks/frameworks.
Finally, we can see that \textit{with the pre-trained model, fine-tuning with only 10--20\% of data (the two leftmost blue circles) generates comparative performance to the supervised learning with all 100\% of training data (the rightmost purple diamond), demonstrating the superiority of GNN pre-training, especially when the label is scarce.}

\vpara{Results for homogeneous graphs.}In addition to heterogeneous graphs, we also test whether the \method pre-training framework can benefit downstream tasks on homogeneous graphs. 
Specifically, we pre-train and fine-tune on two homogeneous graphs: 1) the paper citation network extracted from the field of CS in OAG, on which the topic of each paper is predicted; 2) the Reddit network consisting of Reddit posts, on which the community of each post is inferred.
As there is only one type of nodes and edges in homogeneous graphs, we  require one single set of edge and attribute decoders for pre-training. 
HGT is used as the base pre-training model by ignoring its heterogeneous components.  
The fine-tuned results with 10\% labeled data are presented in Table~\ref{tab:homo}. 
We can observe that the downstream tasks on both homogeneous graphs can benefit from all pre-training frameworks, among which the proposed \method offers the largest performance gains. 


\section{Conclusion}\label{sec:conclusion}

In this work, we study the problem of graph neural network pre-training. 
We present \method---a  
generative GNN pre-training framework. 
In \method, we design the graph generation factorization to guide the base GNN model to autoregressively reconstruct both the attributes and structure of the input graph. 
Furthermore, we propose to separate the attribute and edge generation nodes to avoid information leakage. 
In addition, we introduce the adaptive node representation queue to mitigate the gap between the likelihoods over the sampled graph and the full graph. 
The pre-trained GNNs with fine-tuning over few labeled data can achieve significant performance gains on various downstream tasks across different datasets. 
Also,  \method  is robust to different transfer settings between pre-training and fine-tuning. 
Finally, we find that fine-tuning the generative pre-trained GNN model with 10--20\% of labeled data offers comparative performance for downstream tasks to the supervised GNN model with 100\% of training data.

\vpara{Acknowledgements.} This work is partially supported by NSF III-1705169, NSF CAREER Award 1741634, NSF 1937599, DARPA HR00112090027, DARPA N660011924032, Okawa Foundation Grant, and Amazon Research Award.

\bibliographystyle{ACM-Reference-Format}\balance
\bibliography{kdd}


\begin{thebibliography}{45}


\ifx \showCODEN    \undefined \def \showCODEN     #1{\unskip}     \fi
\ifx \showDOI      \undefined \def \showDOI       #1{#1}\fi
\ifx \showISBNx    \undefined \def \showISBNx     #1{\unskip}     \fi
\ifx \showISBNxiii \undefined \def \showISBNxiii  #1{\unskip}     \fi
\ifx \showISSN     \undefined \def \showISSN      #1{\unskip}     \fi
\ifx \showLCCN     \undefined \def \showLCCN      #1{\unskip}     \fi
\ifx \shownote     \undefined \def \shownote      #1{#1}          \fi
\ifx \showarticletitle \undefined \def \showarticletitle #1{#1}   \fi
\ifx \showURL      \undefined \def \showURL       {\relax}        \fi
\providecommand\bibfield[2]{#2}
\providecommand\bibinfo[2]{#2}
\providecommand\natexlab[1]{#1}
\providecommand\showeprint[2][]{arXiv:#2}

\bibitem[\protect\citeauthoryear{Bruna, Zaremba, Szlam, and LeCun}{Bruna
  et~al\mbox{.}}{2013}]%
        {bruna2013spectral}
\bibfield{author}{\bibinfo{person}{Joan Bruna}, \bibinfo{person}{Wojciech
  Zaremba}, \bibinfo{person}{Arthur Szlam}, {and} \bibinfo{person}{Yann
  LeCun}.} \bibinfo{year}{2013}\natexlab{}.
\newblock \showarticletitle{Spectral networks and locally connected networks on
  graphs}.
\newblock \bibinfo{journal}{\emph{arXiv:1312.6203}} (\bibinfo{year}{2013}).
\newblock


\bibitem[\protect\citeauthoryear{Chen, Kornblith, Norouzi, and Hinton}{Chen
  et~al\mbox{.}}{2020}]%
        {DBLP:journals/corr/abs-2002-05709}
\bibfield{author}{\bibinfo{person}{Ting Chen}, \bibinfo{person}{Simon
  Kornblith}, \bibinfo{person}{Mohammad Norouzi}, {and}
  \bibinfo{person}{Geoffrey~E. Hinton}.} \bibinfo{year}{2020}\natexlab{}.
\newblock \showarticletitle{A Simple Framework for Contrastive Learning of
  Visual Representations}.
\newblock \bibinfo{journal}{\emph{arxiv:2002.05709}} (\bibinfo{year}{2020}).
\newblock


\bibitem[\protect\citeauthoryear{Deng, Dong, Socher, Li, Li, and Fei-Fei}{Deng
  et~al\mbox{.}}{2009}]%
        {deng2009imagenet}
\bibfield{author}{\bibinfo{person}{Jia Deng}, \bibinfo{person}{Wei Dong},
  \bibinfo{person}{Richard Socher}, \bibinfo{person}{Li-Jia Li},
  \bibinfo{person}{Kai Li}, {and} \bibinfo{person}{Li Fei-Fei}.}
  \bibinfo{year}{2009}\natexlab{}.
\newblock \showarticletitle{Imagenet: A large-scale hierarchical image
  database}.
\newblock  (\bibinfo{year}{2009}).
\newblock


\bibitem[\protect\citeauthoryear{Devlin, Chang, Lee, and Toutanova}{Devlin
  et~al\mbox{.}}{2019}]%
        {DBLP:journals/corr/abs-1810-04805}
\bibfield{author}{\bibinfo{person}{Jacob Devlin}, \bibinfo{person}{Ming{-}Wei
  Chang}, \bibinfo{person}{Kenton Lee}, {and} \bibinfo{person}{Kristina
  Toutanova}.} \bibinfo{year}{2019}\natexlab{}.
\newblock \showarticletitle{{BERT:} Pre-training of Deep Bidirectional
  Transformers for Language Understanding}. In
  \bibinfo{booktitle}{\emph{{NAACL} 2019}}.
\newblock


\bibitem[\protect\citeauthoryear{Donahue, Jia, Vinyals, Hoffman, Zhang, Tzeng,
  and Darrell}{Donahue et~al\mbox{.}}{2014}]%
        {donahue2014decaf}
\bibfield{author}{\bibinfo{person}{Jeff Donahue}, \bibinfo{person}{Yangqing
  Jia}, \bibinfo{person}{Oriol Vinyals}, \bibinfo{person}{Judy Hoffman},
  \bibinfo{person}{Ning Zhang}, \bibinfo{person}{Eric Tzeng}, {and}
  \bibinfo{person}{Trevor Darrell}.} \bibinfo{year}{2014}\natexlab{}.
\newblock \showarticletitle{Decaf: A deep convolutional activation feature for
  generic visual recognition}. In \bibinfo{booktitle}{\emph{{ICML} 2014}}.
\newblock


\bibitem[\protect\citeauthoryear{Dong, Chawla, and Swami}{Dong
  et~al\mbox{.}}{2017}]%
        {dong2017metapath2vec}
\bibfield{author}{\bibinfo{person}{Yuxiao Dong}, \bibinfo{person}{Nitesh~V
  Chawla}, {and} \bibinfo{person}{Ananthram Swami}.}
  \bibinfo{year}{2017}\natexlab{}.
\newblock \showarticletitle{metapath2vec: Scalable Representation Learning for
  Heterogeneous Networks}. In \bibinfo{booktitle}{\emph{KDD 2017}}.
\newblock


\bibitem[\protect\citeauthoryear{Dong, Hu, Wang, Sun, and Tang}{Dong
  et~al\mbox{.}}{2020}]%
        {dong2020HeterNRL}
\bibfield{author}{\bibinfo{person}{Yuxiao Dong}, \bibinfo{person}{Ziniu Hu},
  \bibinfo{person}{Kuansan Wang}, \bibinfo{person}{Yizhou Sun}, {and}
  \bibinfo{person}{Jie Tang}.} \bibinfo{year}{2020}\natexlab{}.
\newblock \showarticletitle{Heterogeneous Network Representation Learning}. In
  \bibinfo{booktitle}{\emph{IJCAI}}.
\newblock


\bibitem[\protect\citeauthoryear{Fey and Lenssen}{Fey and Lenssen}{2019}]%
        {pyG}
\bibfield{author}{\bibinfo{person}{Matthias Fey} {and}
  \bibinfo{person}{Jan~Eric Lenssen}.} \bibinfo{year}{2019}\natexlab{}.
\newblock \showarticletitle{Fast Graph Representation Learning with PyTorch
  Geometric}.
\newblock \bibinfo{journal}{\emph{ICLR Workshop}} (\bibinfo{year}{2019}).
\newblock


\bibitem[\protect\citeauthoryear{Gilmer, Schoenholz, Riley, Vinyals, and
  Dahl}{Gilmer et~al\mbox{.}}{2017}]%
        {gilmer2017neural}
\bibfield{author}{\bibinfo{person}{Justin Gilmer}, \bibinfo{person}{Samuel~S.
  Schoenholz}, \bibinfo{person}{Patrick~F. Riley}, \bibinfo{person}{Oriol
  Vinyals}, {and} \bibinfo{person}{George~E. Dahl}.}
  \bibinfo{year}{2017}\natexlab{}.
\newblock \showarticletitle{Neural Message Passing for Quantum Chemistry}. In
  \bibinfo{booktitle}{\emph{{ICML} 2017}}.
\newblock


\bibitem[\protect\citeauthoryear{Girshick, Donahue, Darrell, and
  Malik}{Girshick et~al\mbox{.}}{2014}]%
        {girshick2014rich}
\bibfield{author}{\bibinfo{person}{Ross Girshick}, \bibinfo{person}{Jeff
  Donahue}, \bibinfo{person}{Trevor Darrell}, {and} \bibinfo{person}{Jitendra
  Malik}.} \bibinfo{year}{2014}\natexlab{}.
\newblock \showarticletitle{Rich feature hierarchies for accurate object
  detection and semantic segmentation}. In \bibinfo{booktitle}{\emph{{CVPR}
  2014}}.
\newblock


\bibitem[\protect\citeauthoryear{Grover and Leskovec}{Grover and
  Leskovec}{2016}]%
        {grover2016node2vec}
\bibfield{author}{\bibinfo{person}{Aditya Grover} {and} \bibinfo{person}{Jure
  Leskovec}.} \bibinfo{year}{2016}\natexlab{}.
\newblock \showarticletitle{node2vec: Scalable feature learning for networks}.
  In \bibinfo{booktitle}{\emph{{KDD} 2016}}.
\newblock


\bibitem[\protect\citeauthoryear{Hamilton, Ying, and Leskovec}{Hamilton
  et~al\mbox{.}}{2017}]%
        {graphsage}
\bibfield{author}{\bibinfo{person}{William~L. Hamilton},
  \bibinfo{person}{Zhitao Ying}, {and} \bibinfo{person}{Jure Leskovec}.}
  \bibinfo{year}{2017}\natexlab{}.
\newblock \showarticletitle{Inductive Representation Learning on Large Graphs}.
  In \bibinfo{booktitle}{\emph{{NeurIPS} 2017}}.
\newblock


\bibitem[\protect\citeauthoryear{He, Fan, Wu, Xie, and Girshick}{He
  et~al\mbox{.}}{2019}]%
        {he2019momentum}
\bibfield{author}{\bibinfo{person}{Kaiming He}, \bibinfo{person}{Haoqi Fan},
  \bibinfo{person}{Yuxin Wu}, \bibinfo{person}{Saining Xie}, {and}
  \bibinfo{person}{Ross Girshick}.} \bibinfo{year}{2019}\natexlab{}.
\newblock \showarticletitle{Momentum contrast for unsupervised visual
  representation learning}.
\newblock \bibinfo{journal}{\emph{arXiv:1911.05722}} (\bibinfo{year}{2019}).
\newblock


\bibitem[\protect\citeauthoryear{Hu, Liu, Gomes, Zitnik, Liang, Pande, and
  Leskovec}{Hu et~al\mbox{.}}{2020b}]%
        {pretrain}
\bibfield{author}{\bibinfo{person}{Weihua Hu}, \bibinfo{person}{Bowen Liu},
  \bibinfo{person}{Joseph Gomes}, \bibinfo{person}{Marinka Zitnik},
  \bibinfo{person}{Percy Liang}, \bibinfo{person}{Vijay~S. Pande}, {and}
  \bibinfo{person}{Jure Leskovec}.} \bibinfo{year}{2020}\natexlab{b}.
\newblock \showarticletitle{Strategies for Pre-training Graph Neural Networks}.
  In \bibinfo{booktitle}{\emph{{ICLR} 2020}}.
\newblock


\bibitem[\protect\citeauthoryear{Hu, Dong, Wang, and Sun}{Hu
  et~al\mbox{.}}{2020a}]%
        {hgt}
\bibfield{author}{\bibinfo{person}{Ziniu Hu}, \bibinfo{person}{Yuxiao Dong},
  \bibinfo{person}{Kuansan Wang}, {and} \bibinfo{person}{Yizhou Sun}.}
  \bibinfo{year}{2020}\natexlab{a}.
\newblock \showarticletitle{Heterogeneous Graph Transformer}. In
  \bibinfo{booktitle}{\emph{{WWW} 2020}}.
\newblock


\bibitem[\protect\citeauthoryear{Kipf and Welling}{Kipf and Welling}{2016}]%
        {DBLP:journals/corr/KipfW16a}
\bibfield{author}{\bibinfo{person}{Thomas~N. Kipf} {and} \bibinfo{person}{Max
  Welling}.} \bibinfo{year}{2016}\natexlab{}.
\newblock \showarticletitle{Variational Graph Auto-Encoders}.
\newblock \bibinfo{journal}{\emph{arXiv:1611.07308}} (\bibinfo{year}{2016}).
\newblock


\bibitem[\protect\citeauthoryear{Kipf and Welling}{Kipf and Welling}{2017}]%
        {gcn}
\bibfield{author}{\bibinfo{person}{Thomas~N. Kipf} {and} \bibinfo{person}{Max
  Welling}.} \bibinfo{year}{2017}\natexlab{}.
\newblock \showarticletitle{Semi-Supervised Classification with Graph
  Convolutional Networks}. In \bibinfo{booktitle}{\emph{{ICLR} 2017}}.
\newblock


\bibitem[\protect\citeauthoryear{Liao, Li, Song, Wang, Hamilton, Duvenaud,
  Urtasun, and Zemel}{Liao et~al\mbox{.}}{2019}]%
        {DBLP:conf/nips/LiaoLSWHDUZ19}
\bibfield{author}{\bibinfo{person}{Renjie Liao}, \bibinfo{person}{Yujia Li},
  \bibinfo{person}{Yang Song}, \bibinfo{person}{Shenlong Wang},
  \bibinfo{person}{William~L. Hamilton}, \bibinfo{person}{David Duvenaud},
  \bibinfo{person}{Raquel Urtasun}, {and} \bibinfo{person}{Richard~S. Zemel}.}
  \bibinfo{year}{2019}\natexlab{}.
\newblock \showarticletitle{Efficient Graph Generation with Graph Recurrent
  Attention Networks}. In \bibinfo{booktitle}{\emph{{NeurIPS} 2019}}.
\newblock


\bibitem[\protect\citeauthoryear{Liu}{Liu}{2011}]%
        {DBLP:books/daglib/0027504}
\bibfield{author}{\bibinfo{person}{Tie{-}Yan Liu}.}
  \bibinfo{year}{2011}\natexlab{}.
\newblock \bibinfo{booktitle}{\emph{Learning to Rank for Information
  Retrieval}}.
\newblock \bibinfo{publisher}{Springer}.
\newblock
\showISBNx{978-3-642-14266-6}


\bibitem[\protect\citeauthoryear{Loshchilov and Hutter}{Loshchilov and
  Hutter}{2017}]%
        {DBLP:conf/iclr/LoshchilovH17}
\bibfield{author}{\bibinfo{person}{Ilya Loshchilov} {and}
  \bibinfo{person}{Frank Hutter}.} \bibinfo{year}{2017}\natexlab{}.
\newblock \showarticletitle{{SGDR:} Stochastic Gradient Descent with Warm
  Restarts}. In \bibinfo{booktitle}{\emph{{ICLR} 2017}}.
\newblock


\bibitem[\protect\citeauthoryear{Loshchilov and Hutter}{Loshchilov and
  Hutter}{2019}]%
        {DBLP:conf/iclr/LoshchilovH19}
\bibfield{author}{\bibinfo{person}{Ilya Loshchilov} {and}
  \bibinfo{person}{Frank Hutter}.} \bibinfo{year}{2019}\natexlab{}.
\newblock \showarticletitle{Decoupled Weight Decay Regularization}. In
  \bibinfo{booktitle}{\emph{{ICLR} 2019}}.
\newblock


\bibitem[\protect\citeauthoryear{Mikolov, Sutskever, Chen, Corrado, and
  Dean}{Mikolov et~al\mbox{.}}{2013}]%
        {mikolov2013distributed}
\bibfield{author}{\bibinfo{person}{Tomas Mikolov}, \bibinfo{person}{Ilya
  Sutskever}, \bibinfo{person}{Kai Chen}, \bibinfo{person}{Greg~S Corrado},
  {and} \bibinfo{person}{Jeff Dean}.} \bibinfo{year}{2013}\natexlab{}.
\newblock \showarticletitle{Distributed representations of words and phrases
  and their compositionality}. In \bibinfo{booktitle}{\emph{{NIPS} 2013}}.
\newblock


\bibitem[\protect\citeauthoryear{Ni, Li, and McAuley}{Ni et~al\mbox{.}}{2019}]%
        {DBLP:conf/emnlp/NiLM19}
\bibfield{author}{\bibinfo{person}{Jianmo Ni}, \bibinfo{person}{Jiacheng Li},
  {and} \bibinfo{person}{Julian~J. McAuley}.} \bibinfo{year}{2019}\natexlab{}.
\newblock \showarticletitle{Justifying Recommendations using Distantly-Labeled
  Reviews and Fine-Grained Aspects}. In \bibinfo{booktitle}{\emph{{EMNLP}
  2019}}.
\newblock


\bibitem[\protect\citeauthoryear{Pathak, Kr{\"{a}}henb{\"{u}}hl, Donahue,
  Darrell, and Efros}{Pathak et~al\mbox{.}}{2016}]%
        {DBLP:conf/cvpr/PathakKDDE16}
\bibfield{author}{\bibinfo{person}{Deepak Pathak}, \bibinfo{person}{Philipp
  Kr{\"{a}}henb{\"{u}}hl}, \bibinfo{person}{Jeff Donahue},
  \bibinfo{person}{Trevor Darrell}, {and} \bibinfo{person}{Alexei~A. Efros}.}
  \bibinfo{year}{2016}\natexlab{}.
\newblock \showarticletitle{Context Encoders: Feature Learning by Inpainting}.
  In \bibinfo{booktitle}{\emph{{CVPR} 2016}}.
\newblock


\bibitem[\protect\citeauthoryear{Pennington, Socher, and Manning}{Pennington
  et~al\mbox{.}}{2014}]%
        {pennington2014glove}
\bibfield{author}{\bibinfo{person}{Jeffrey Pennington},
  \bibinfo{person}{Richard Socher}, {and} \bibinfo{person}{Christopher
  Manning}.} \bibinfo{year}{2014}\natexlab{}.
\newblock \showarticletitle{Glove: Global vectors for word representation}. In
  \bibinfo{booktitle}{\emph{{EMNLP} 2014}}.
\newblock


\bibitem[\protect\citeauthoryear{Qiu, Dong, Ma, Li, Wang, and Tang}{Qiu
  et~al\mbox{.}}{2018}]%
        {qiu2018network}
\bibfield{author}{\bibinfo{person}{Jiezhong Qiu}, \bibinfo{person}{Yuxiao
  Dong}, \bibinfo{person}{Hao Ma}, \bibinfo{person}{Jian Li},
  \bibinfo{person}{Kuansan Wang}, {and} \bibinfo{person}{Jie Tang}.}
  \bibinfo{year}{2018}\natexlab{}.
\newblock \showarticletitle{Network embedding as matrix factorization: Unifying
  deepwalk, line, pte, and node2vec}. In \bibinfo{booktitle}{\emph{WSDM '18}}.
  \bibinfo{pages}{459--467}.
\newblock


\bibitem[\protect\citeauthoryear{Radford, Wu, Child, Luan, Amodei, and
  Sutskever}{Radford et~al\mbox{.}}{2019}]%
        {radford2019language}
\bibfield{author}{\bibinfo{person}{Alec Radford}, \bibinfo{person}{Jeff Wu},
  \bibinfo{person}{Rewon Child}, \bibinfo{person}{David Luan},
  \bibinfo{person}{Dario Amodei}, {and} \bibinfo{person}{Ilya Sutskever}.}
  \bibinfo{year}{2019}\natexlab{}.
\newblock \showarticletitle{Language Models are Unsupervised Multitask
  Learners}.
\newblock  (\bibinfo{year}{2019}).
\newblock


\bibitem[\protect\citeauthoryear{Schlichtkrull, Kipf, Bloem, van~den Berg,
  Titov, and Welling}{Schlichtkrull et~al\mbox{.}}{2018}]%
        {DBLP:conf/esws/SchlichtkrullKB18}
\bibfield{author}{\bibinfo{person}{Michael~Sejr Schlichtkrull},
  \bibinfo{person}{Thomas~N. Kipf}, \bibinfo{person}{Peter Bloem},
  \bibinfo{person}{Rianne van~den Berg}, \bibinfo{person}{Ivan Titov}, {and}
  \bibinfo{person}{Max Welling}.} \bibinfo{year}{2018}\natexlab{}.
\newblock \showarticletitle{Modeling Relational Data with Graph Convolutional
  Networks}. In \bibinfo{booktitle}{\emph{{ESWC} 2018}}.
\newblock


\bibitem[\protect\citeauthoryear{Sun, Hoffmann, Verma, and Tang}{Sun
  et~al\mbox{.}}{2020}]%
        {infograph}
\bibfield{author}{\bibinfo{person}{Fan-Yun Sun}, \bibinfo{person}{Jordan
  Hoffmann}, \bibinfo{person}{Vikas Verma}, {and} \bibinfo{person}{Jian Tang}.}
  \bibinfo{year}{2020}\natexlab{}.
\newblock \showarticletitle{InfoGraph: Unsupervised and Semi-supervised
  Graph-Level Representation Learning via Mutual Information Maximization}. In
  \bibinfo{booktitle}{\emph{{ICLR} 2020}}.
\newblock


\bibitem[\protect\citeauthoryear{Sun and Han}{Sun and Han}{2012}]%
        {Sun:BOOK2012}
\bibfield{author}{\bibinfo{person}{Yizhou Sun} {and} \bibinfo{person}{Jiawei
  Han}.} \bibinfo{year}{2012}\natexlab{}.
\newblock \bibinfo{booktitle}{\emph{Mining Heterogeneous Information Networks:
  Principles and Methodologies}}.
\newblock \bibinfo{publisher}{Morgan \& Claypool Publishers}.
\newblock


\bibitem[\protect\citeauthoryear{Sun, Han, Yan, Yu, and Wu}{Sun
  et~al\mbox{.}}{2011}]%
        {Sun:VLDB11}
\bibfield{author}{\bibinfo{person}{Yizhou Sun}, \bibinfo{person}{Jiawei Han},
  \bibinfo{person}{Xifeng Yan}, \bibinfo{person}{Philip~S. Yu}, {and}
  \bibinfo{person}{Tianyi Wu}.} \bibinfo{year}{2011}\natexlab{}.
\newblock \showarticletitle{Pathsim: Meta path-based top-k similarity search in
  heterogeneous information networks}. In \bibinfo{booktitle}{\emph{{VLDB}
  2011}}.
\newblock


\bibitem[\protect\citeauthoryear{Sun, Norick, Han, Yan, Yu, and Yu}{Sun
  et~al\mbox{.}}{2012}]%
        {DBLP:conf/kdd/SunNHYYY12}
\bibfield{author}{\bibinfo{person}{Yizhou Sun}, \bibinfo{person}{Brandon
  Norick}, \bibinfo{person}{Jiawei Han}, \bibinfo{person}{Xifeng Yan},
  \bibinfo{person}{Philip~S. Yu}, {and} \bibinfo{person}{Xiao Yu}.}
  \bibinfo{year}{2012}\natexlab{}.
\newblock \showarticletitle{Integrating meta-path selection with user-guided
  object clustering in heterogeneous information networks}. In
  \bibinfo{booktitle}{\emph{{KDD} 2012}}.
\newblock


\bibitem[\protect\citeauthoryear{Tang, Qu, Wang, Zhang, Yan, and Mei}{Tang
  et~al\mbox{.}}{2015}]%
        {tang2015line}
\bibfield{author}{\bibinfo{person}{Jian Tang}, \bibinfo{person}{Meng Qu},
  \bibinfo{person}{Mingzhe Wang}, \bibinfo{person}{Ming Zhang},
  \bibinfo{person}{Jun Yan}, {and} \bibinfo{person}{Qiaozhu Mei}.}
  \bibinfo{year}{2015}\natexlab{}.
\newblock \showarticletitle{Line: Large-scale information network embedding}.
  In \bibinfo{booktitle}{\emph{{WWW} 2015}}.
\newblock


\bibitem[\protect\citeauthoryear{Tang, Zhang, Yao, Li, Zhang, and Su}{Tang
  et~al\mbox{.}}{2008}]%
        {tang2008arnetminer}
\bibfield{author}{\bibinfo{person}{Jie Tang}, \bibinfo{person}{Jing Zhang},
  \bibinfo{person}{Limin Yao}, \bibinfo{person}{Juanzi Li}, \bibinfo{person}{Li
  Zhang}, {and} \bibinfo{person}{Zhong Su}.} \bibinfo{year}{2008}\natexlab{}.
\newblock \showarticletitle{Arnetminer: extraction and mining of academic
  social networks}. In \bibinfo{booktitle}{\emph{KDD 2008}}.
\newblock


\bibitem[\protect\citeauthoryear{van~den Oord, Li, and Vinyals}{van~den Oord
  et~al\mbox{.}}{2018}]%
        {DBLP:journals/corr/abs-1807-03748}
\bibfield{author}{\bibinfo{person}{A{\"{a}}ron van~den Oord},
  \bibinfo{person}{Yazhe Li}, {and} \bibinfo{person}{Oriol Vinyals}.}
  \bibinfo{year}{2018}\natexlab{}.
\newblock \showarticletitle{Representation Learning with Contrastive Predictive
  Coding}.
\newblock \bibinfo{journal}{\emph{arXiv:1807.03748}} (\bibinfo{year}{2018}).
\newblock


\bibitem[\protect\citeauthoryear{Velickovic, Cucurull, Casanova, Romero,
  Li{\`{o}}, and Bengio}{Velickovic et~al\mbox{.}}{2018}]%
        {gat}
\bibfield{author}{\bibinfo{person}{Petar Velickovic}, \bibinfo{person}{Guillem
  Cucurull}, \bibinfo{person}{Arantxa Casanova}, \bibinfo{person}{Adriana
  Romero}, \bibinfo{person}{Pietro Li{\`{o}}}, {and} \bibinfo{person}{Yoshua
  Bengio}.} \bibinfo{year}{2018}\natexlab{}.
\newblock \showarticletitle{Graph Attention Networks}. In
  \bibinfo{booktitle}{\emph{{ICLR} 2018}}.
\newblock


\bibitem[\protect\citeauthoryear{Velickovic, Fedus, Hamilton, Li{\`{o}},
  Bengio, and Hjelm}{Velickovic et~al\mbox{.}}{2019}]%
        {DBLP:journals/corr/abs-1809-10341}
\bibfield{author}{\bibinfo{person}{Petar Velickovic}, \bibinfo{person}{William
  Fedus}, \bibinfo{person}{William~L. Hamilton}, \bibinfo{person}{Pietro
  Li{\`{o}}}, \bibinfo{person}{Yoshua Bengio}, {and} \bibinfo{person}{R.~Devon
  Hjelm}.} \bibinfo{year}{2019}\natexlab{}.
\newblock \showarticletitle{Deep Graph Infomax}. In
  \bibinfo{booktitle}{\emph{{ICLR} 2019}}.
\newblock


\bibitem[\protect\citeauthoryear{{Wang}, {Shen}, {Huang}, {Wu}, {Dong}, and
  {Kanakia}}{{Wang} et~al\mbox{.}}{2020}]%
        {wang2020mag}
\bibfield{author}{\bibinfo{person}{Kuansan {Wang}}, \bibinfo{person}{Zhihong
  {Shen}}, \bibinfo{person}{Chiyuan {Huang}}, \bibinfo{person}{Chieh-Han {Wu}},
  \bibinfo{person}{Yuxiao {Dong}}, {and} \bibinfo{person}{Anshul {Kanakia}}.}
  \bibinfo{year}{2020}\natexlab{}.
\newblock \showarticletitle{Microsoft Academic Graph: When experts are not
  enough}.
\newblock \bibinfo{journal}{\emph{Quantitative Science Studies}}
  \bibinfo{volume}{1}, \bibinfo{number}{1} (\bibinfo{year}{2020}),
  \bibinfo{pages}{396--413}.
\newblock


\bibitem[\protect\citeauthoryear{Wang, Ji, Shi, Wang, Ye, Cui, and Yu}{Wang
  et~al\mbox{.}}{2019}]%
        {DBLP:conf/www/WangJSWYCY19}
\bibfield{author}{\bibinfo{person}{Xiao Wang}, \bibinfo{person}{Houye Ji},
  \bibinfo{person}{Chuan Shi}, \bibinfo{person}{Bai Wang},
  \bibinfo{person}{Yanfang Ye}, \bibinfo{person}{Peng Cui}, {and}
  \bibinfo{person}{Philip~S. Yu}.} \bibinfo{year}{2019}\natexlab{}.
\newblock \showarticletitle{Heterogeneous Graph Attention Network}. In
  \bibinfo{booktitle}{\emph{{WWW} 2019}}.
\newblock


\bibitem[\protect\citeauthoryear{Wolf, Debut, Sanh, Chaumond, Delangue, Moi,
  Cistac, Rault, Louf, Funtowicz, and Brew}{Wolf et~al\mbox{.}}{2019}]%
        {wolf2019transformers}
\bibfield{author}{\bibinfo{person}{Thomas Wolf}, \bibinfo{person}{Lysandre
  Debut}, \bibinfo{person}{Victor Sanh}, \bibinfo{person}{Julien Chaumond},
  \bibinfo{person}{Clement Delangue}, \bibinfo{person}{Anthony Moi},
  \bibinfo{person}{Pierric Cistac}, \bibinfo{person}{Tim Rault},
  \bibinfo{person}{Rémi Louf}, \bibinfo{person}{Morgan Funtowicz}, {and}
  \bibinfo{person}{Jamie Brew}.} \bibinfo{year}{2019}\natexlab{}.
\newblock \bibinfo{title}{Transformers: State-of-the-art Natural Language
  Processing}.
\newblock
\newblock
\showeprint[arxiv]{cs.CL/1910.03771}


\bibitem[\protect\citeauthoryear{Yang, Dai, Yang, Carbonell, Salakhutdinov, and
  Le}{Yang et~al\mbox{.}}{2019}]%
        {xlnet}
\bibfield{author}{\bibinfo{person}{Zhilin Yang}, \bibinfo{person}{Zihang Dai},
  \bibinfo{person}{Yiming Yang}, \bibinfo{person}{Jaime~G. Carbonell},
  \bibinfo{person}{Ruslan Salakhutdinov}, {and} \bibinfo{person}{Quoc~V. Le}.}
  \bibinfo{year}{2019}\natexlab{}.
\newblock \showarticletitle{XLNet: Generalized Autoregressive Pretraining for
  Language Understanding}. In \bibinfo{booktitle}{\emph{{NeurIPS} 2019}}.
\newblock


\bibitem[\protect\citeauthoryear{Ying, He, Chen, Eksombatchai, Hamilton, and
  Leskovec}{Ying et~al\mbox{.}}{2018}]%
        {DBLP:conf/kdd/YingHCEHL18}
\bibfield{author}{\bibinfo{person}{Rex Ying}, \bibinfo{person}{Ruining He},
  \bibinfo{person}{Kaifeng Chen}, \bibinfo{person}{Pong Eksombatchai},
  \bibinfo{person}{William~L. Hamilton}, {and} \bibinfo{person}{Jure
  Leskovec}.} \bibinfo{year}{2018}\natexlab{}.
\newblock \showarticletitle{Graph Convolutional Neural Networks for Web-Scale
  Recommender Systems}. In \bibinfo{booktitle}{\emph{{KDD} 2018}}.
\newblock


\bibitem[\protect\citeauthoryear{You, Ying, Ren, Hamilton, and Leskovec}{You
  et~al\mbox{.}}{2018}]%
        {DBLP:conf/icml/YouYRHL18}
\bibfield{author}{\bibinfo{person}{Jiaxuan You}, \bibinfo{person}{Rex Ying},
  \bibinfo{person}{Xiang Ren}, \bibinfo{person}{William~L. Hamilton}, {and}
  \bibinfo{person}{Jure Leskovec}.} \bibinfo{year}{2018}\natexlab{}.
\newblock \showarticletitle{GraphRNN: Generating Realistic Graphs with Deep
  Auto-regressive Models}. In \bibinfo{booktitle}{\emph{{ICML} 2018}}.
\newblock


\bibitem[\protect\citeauthoryear{Zhang, Liu, Tang, Dong, Yao, Zhang, Gu, Wang,
  Shao, Li, and Wang}{Zhang et~al\mbox{.}}{2019}]%
        {DBLP:conf/kdd/ZhangLTDYZGWSLW19}
\bibfield{author}{\bibinfo{person}{Fanjin Zhang}, \bibinfo{person}{Xiao Liu},
  \bibinfo{person}{Jie Tang}, \bibinfo{person}{Yuxiao Dong},
  \bibinfo{person}{Peiran Yao}, \bibinfo{person}{Jie Zhang},
  \bibinfo{person}{Xiaotao Gu}, \bibinfo{person}{Yan Wang},
  \bibinfo{person}{Bin Shao}, \bibinfo{person}{Rui Li}, {and}
  \bibinfo{person}{Kuansan Wang}.} \bibinfo{year}{2019}\natexlab{}.
\newblock \showarticletitle{{OAG:} Toward Linking Large-scale Heterogeneous
  Entity Graphs}. In \bibinfo{booktitle}{\emph{{KDD} 2019}}.
\newblock


\bibitem[\protect\citeauthoryear{Zou, Hu, Wang, Jiang, Sun, and Gu}{Zou
  et~al\mbox{.}}{2019}]%
        {ladies}
\bibfield{author}{\bibinfo{person}{Difan Zou}, \bibinfo{person}{Ziniu Hu},
  \bibinfo{person}{Yewen Wang}, \bibinfo{person}{Song Jiang},
  \bibinfo{person}{Yizhou Sun}, {and} \bibinfo{person}{Quanquan Gu}.}
  \bibinfo{year}{2019}\natexlab{}.
\newblock \showarticletitle{Layer-Dependent Importance Sampling for Training
  Deep and Large Graph Convolutional Networks}. In
  \bibinfo{booktitle}{\emph{{NeurIPS} 2019}}.
\newblock


\end{thebibliography}

\newpage
\appendix
\begin{figure*}[t!]
    \centering
    \includegraphics[width=1.0\textwidth, trim = 0 0 10 0, clip]{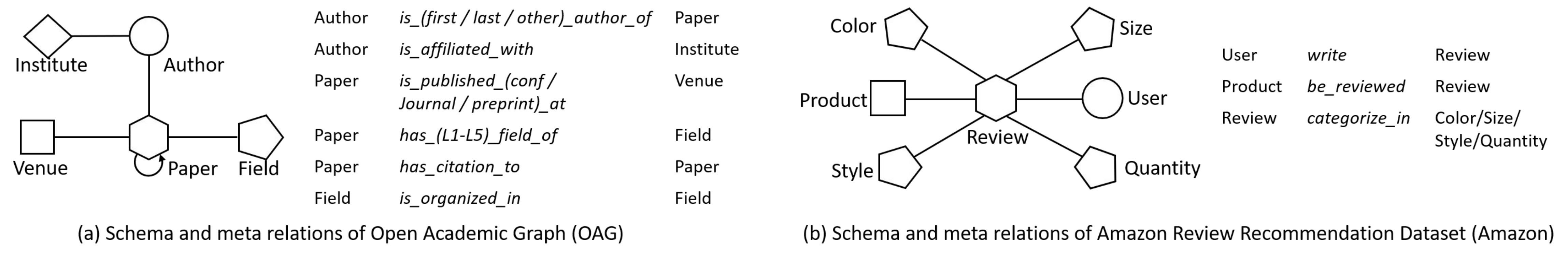}
    \caption{The schema and meta relations of Open Academic Graph and Amazon Review Recommendation Dataset.}
    \label{fig:schema}
\end{figure*}

\begin{table*}[!t]
\centering
\small
\begin{tabular}{ll}
\toprule
Predict Paper Title & Groundtruth Paper Title \\ \midrule
person recognition system using automatic probabilistic classification & person re-identification by probabilistic relative distance comparison
\\ 
a novel framework using spectrum sensing in wireless systems   & a secure collaborative spectrum sensing strategy in cyber physical systems\\ 
a efficient evaluation of a distributed data storage service storage  &   an empirical analysis of a large scale mobile cloud storage service\\ 
parameter control in wireless sensor networks networks networks  & optimal parameter estimation under controlled communication over sensor networks\\
a experimental system for for to the analysis of graphics &  an interactive computer graphics approach to surface representation\\
\bottomrule
\end{tabular} 
\caption{Generated paper title samples. The left column is generated by \method, and the right column is the groundtruth.} 
\label{tab:gen}
\end{table*}

\begin{figure*}[t!]
    \subfigure[Data Percentage: 10\%]{
        \label{fig:d_s}
        \includegraphics[width=0.3\textwidth]{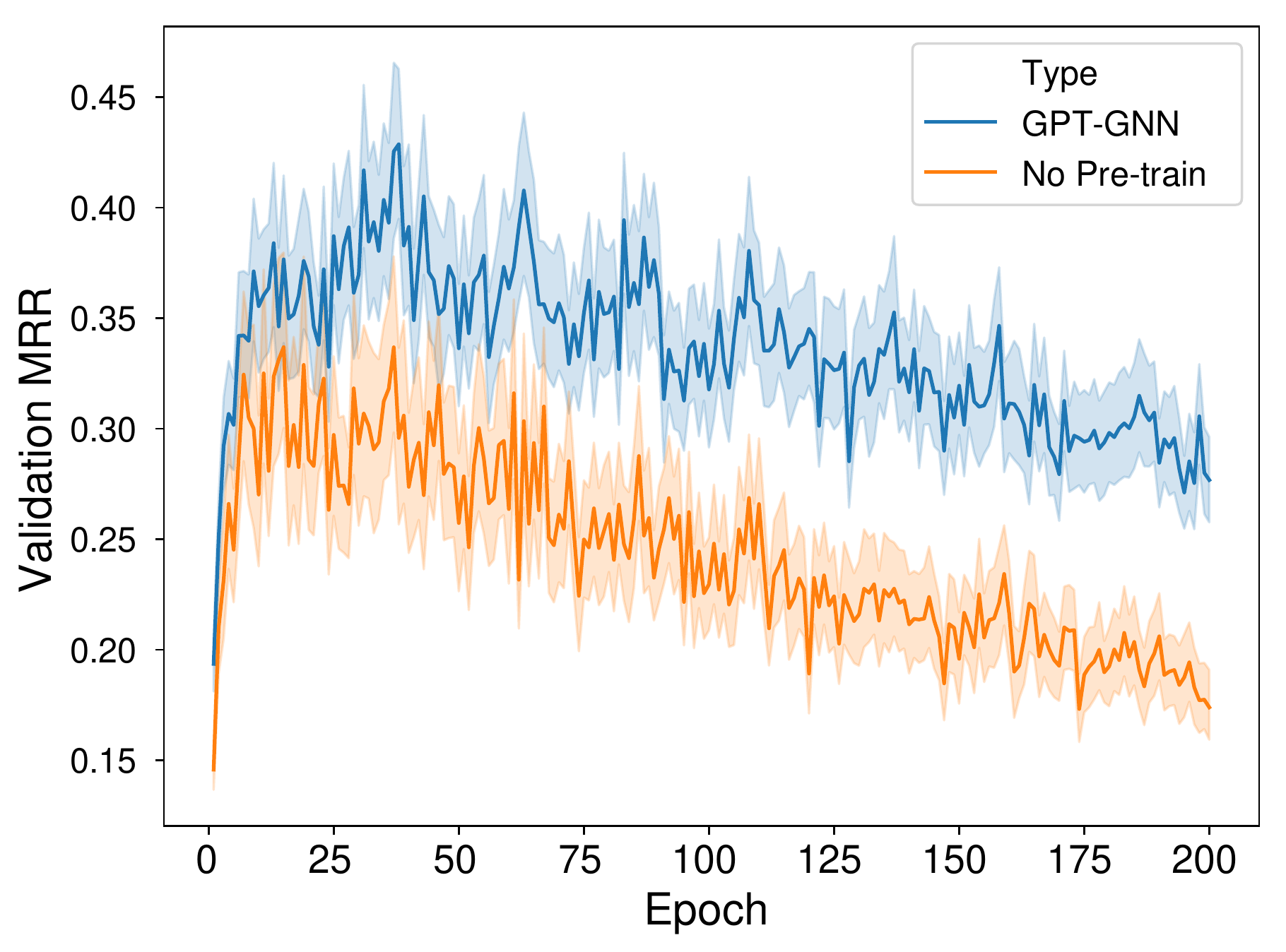}
    }
    \subfigure[Data Percentage: 20\%]{
        \label{fig:d_u}
        \includegraphics[width=0.3\textwidth]{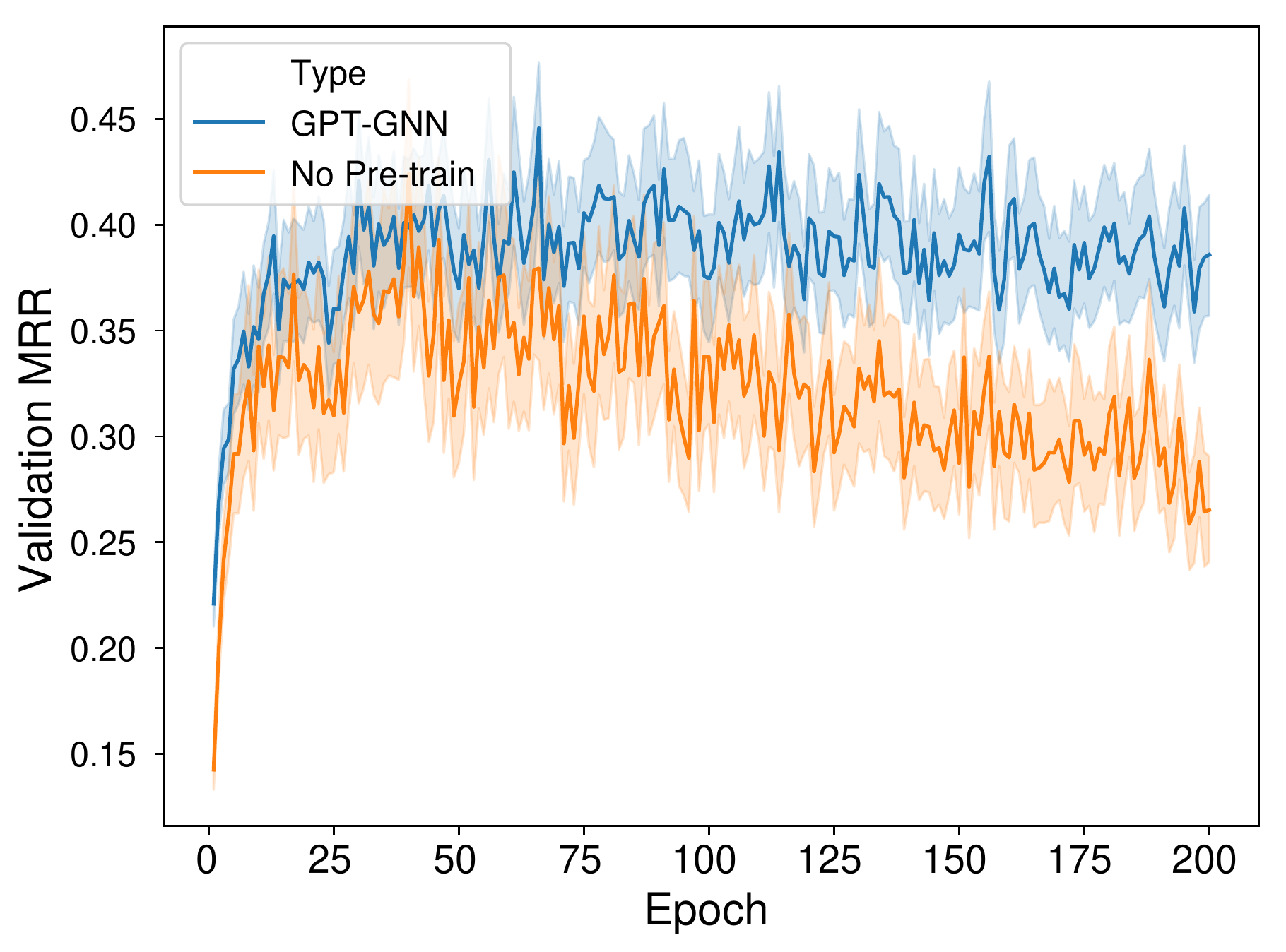}
    }
    \subfigure[Data Percentage: 50\%]{
        \label{fig:s_s}
        \includegraphics[width=0.3\textwidth]{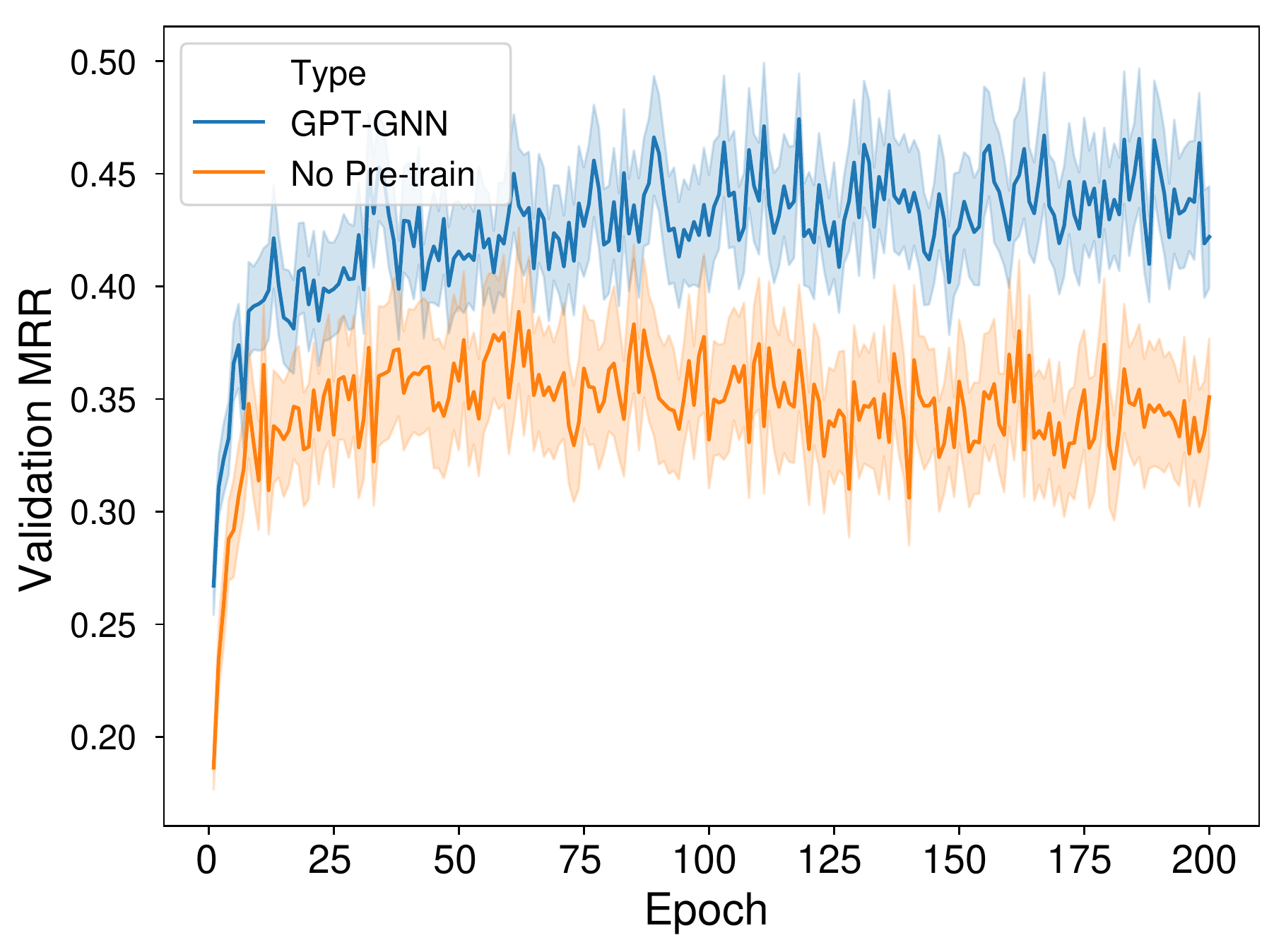}
    }
    \caption{Fine-tuning convergence comparison of GPT-GNN with no-pretrain, under different training data percentage.}
    \label{fig:cv2}
\end{figure*}

\begin{figure}[t!]
    \centering
    \includegraphics[width=0.44\textwidth]{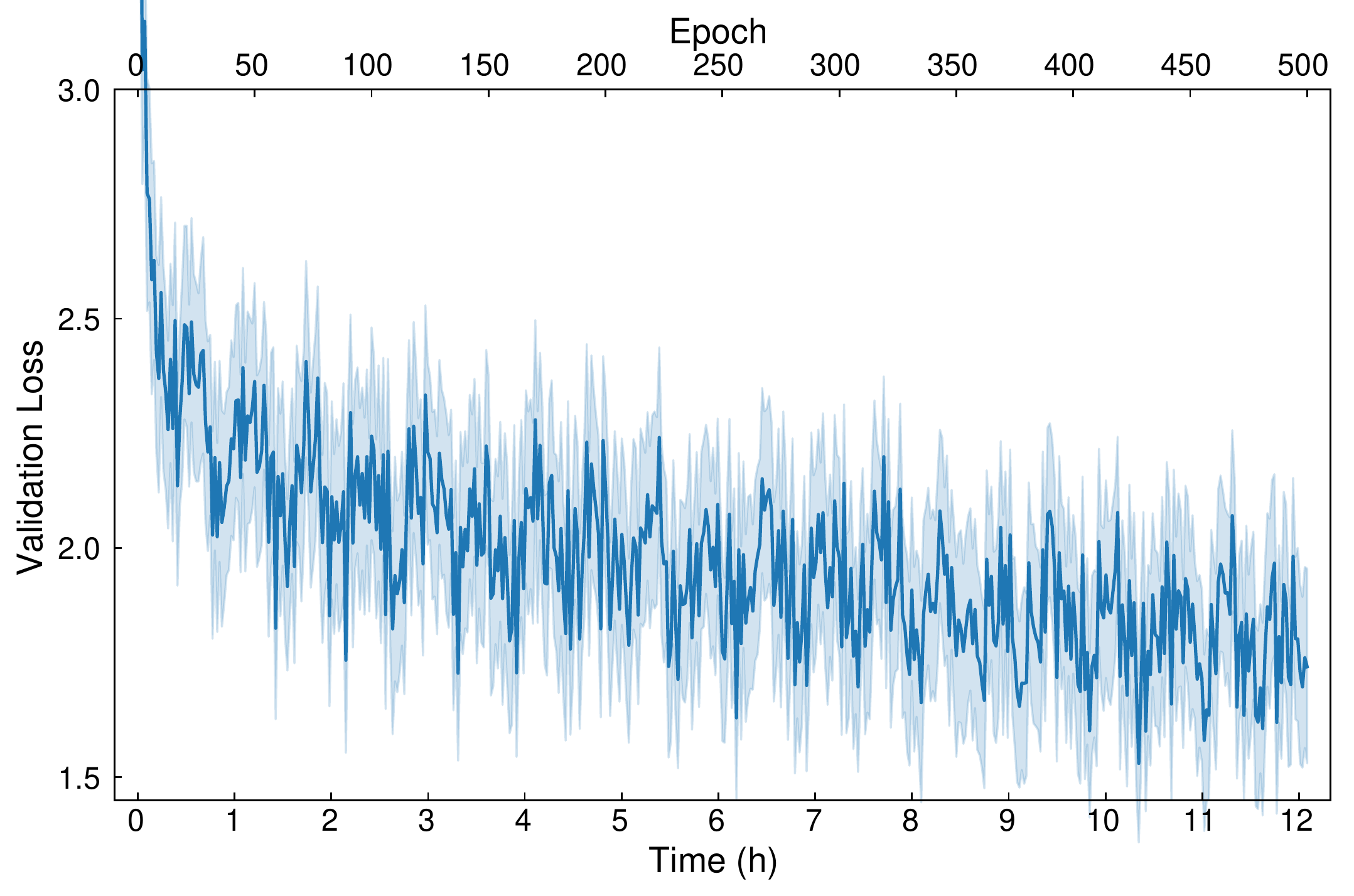}
    \caption{The pre-training convergence curve of GPT-GNN on OAG. It took about 10 hours (400 epochs) to converge.}
    \label{fig:cv1}
\end{figure}

\section{Dataset Details}

We mainly use \textit{Open Academic Graph (OAG)} and \textit{Amazon Review Recommendation Dataset (Amazon)} for evaluation. Both are widely used heterogeneous graph~\cite{Sun:VLDB11, DBLP:conf/kdd/SunNHYYY12, Sun:BOOK2012}. Here we introduce their statistics, schema and how we prepare the attributes and tasks in detail.



\textit{Open Academic Graph (OAG)}~\cite{DBLP:conf/kdd/ZhangLTDYZGWSLW19,wang2020mag,tang2008arnetminer} consists of five types of nodes: `Paper', `Author', `Field', `Venue', and `Institute', and 14 types of edges between these nodes. The schema and meta relations are illustrated in Figure~\ref{fig:schema}(a). For example, the `Field' nodes in the OAG are categorized into six levels from $L_0$ to $L_5$, which are organized with a hierarchical tree (We use `is\_organized\_in' to represent this hierarchy). 
Therefore, we differentiate the `Paper--Field' edges in the corresponding field levels. 
Besides, we differentiate the different author orders (i.e., the first author, the last one, and others) and venue types (i.e., journal, conference, and preprint) as well. 
Finally, the `Self' type corresponds to the self-loop connection, which is widely added in GNN architectures. Despite `Self' and `CoAuthor' edge relationships, which are symmetric, all other edge types $X$ have a reverse edge type $X^{-1}$. 

For downstream tasks, we choose the following three tasks: the prediction of Paper--Field (L2), Paper--Venue, and Author Disambiguation. In the first two tasks, we give a model a paper and want it to predict the correct fields it belongs to or the venue it is published at. We model these three tasks as node classification problem, where we use GNNs to get the contextual node representation of the paper and use a softmax output layer to get its classification results. 
For author disambiguation, we pick all the authors with the same name, and the papers that link to one of these same-name authors. The task is to conduct link prediction between papers and candidate authors. 
After getting the paper and author node representations from GNNs, we use a Neural Tensor Network to get the probability of each author-paper pair to be linked.

For input attributes of heterogeneous graph, as we don't assume the attribute of each data type belongs to the same distribution, and we are free to use the most appropriate attributes to represent each type of node. For paper and author nodes, the node numbers are extremely large. Therefore, traditional node embedding algorithms are not suitable for extracting attributes for them. 
We, therefore, resort to the paper titles for extracting attributes. For each paper, we get its title text and use a pre-trained XLNet~\cite{xlnet, wolf2019transformers} to get the representation of each word in the title. We then average them weighted by each word's attention to get the title representation for each paper. The initial attribute of each author is simply an average of his/her published papers' embeddings. For field, venue and institute nodes, the node numbers are small and we use the metapath2vec model~\cite{dong2017metapath2vec} to train their node embeddings by reflecting the heterogeneous network structures.

\textit{Amazon Review Recommendation Dataset (Amazon)}~\cite{DBLP:conf/emnlp/NiLM19} consists of three types of nodes, including reviews (ratings and text), users and products, and some other meta-types of the product, including color, size, style and quantity. The schema and meta relations are illustrated in Figure~\ref{fig:schema}(b). Compared to a general user-item bipartite graph, this dataset have review in between, each is associated with text information and a rating from 1 to 5. The reviews are also associated with those product meta-type descriptions. For simplicity, we only consider these review-type link as `categorize\_in' type. Thus, there are totally three types of relations in this graph.

For downstream task, we choose the rating classification for each new review. Since the problem is a node-level multi-class classification, we use GNNs to get the contextual node representation of the review and use a softmax output layer to get 5-class prediction.

For input attributes of Amazon, we also use a pre-trained XLNet to get each review embedding, and the attributes for all the other nodes are simple an average of its associated review's embeddings.


\section{Overall Pipeline of GPT-GNN}
The overall pipeline of GPT-GNN is illustrated by Algorithm~\ref{alg:pipeline}. Given an attributed graph $G$, we each time sample a subgraph $\hat{G}$ as training instance for generative pre-training. The first step is to determine the node permutation order $\pi$. To support parallel training, we want to conduct the forward pass for a single run and get the representation of the whole graph, so that we can simultaneously calculate the loss for each node, instead of processing each node recursively. Therefore, we remove all the edges from nodes with higher order to those with lower order according to $\pi$, which means each node can only receive information from nodes with lower order. In this way, they won't leak the information to the autoregressive generative objective, and thus we only need one single round to get node embeddings of the whole graph, which can directly be utilized to conduct generative pre-training.

Afterwards, we need to determine the edges to be masked out. For each node, we get all its outward edges, randomly select a set of edges to be masked out. This corresponds to line~\ref{alg:step2}. Next, we conduct node separation and get contextualized node embeddings for the whole graph in line~\ref{alg:step3}, which will be utilized to calculate generative loss. line~\ref{alg:step4}-\ref{alg:step5}. For both OAG and Amazon, the main nodes that contain meaningful attributes are paper and review nodes, which both have text feature as input. Thus, we only consider them for Attribute Generation, with a 2-layer LSTM as decoder. Note that in line~\ref{alg:step9}, we prepare the negative samples by aggregating both the unconnected nodes within this sampled graph and the previously calculated embeddings stored in the adaptive queue $Q$. This can mitigate the gap between optimizing over sampled graph with over the whole graph. Finally, we optimize the model and update the adaptaive queue in line ~\ref{alg:step6}-\ref{alg:step7}. Afterwards, we can use the pre-trained model as initialization, to fine-tune on other downstream tasks.



\section{Implementation Details and Convergence Curves}
We use a Tesla K80 to run both pre-training and downstream tasks. For graph sampling, we follow the HGSampling~\cite{hgt, ladies} to sample subgraph over large-scale heterogeneous graph. For each type of node, we sample 128 nodes per layer. We repeat sampling for 6 times for OAG and average sampled nodes in the sub-graph is 3561 nodes. We repeat for 8 times for Amazon, and average sampled nodes is 1478. For each batch, we sample 32 graphs to conduct generative pre-training. During GPU training, we conduct multi-process sampling to prepare the pre-training data. Such CPU-GPU cooperation can help us save the sampling time.

We here illustrate the convergence curves for pre-training and fine-tuning. For pre-training, as illustrated in Figure~\ref{fig:cv1}, we show the pre-training validation error curve with respect to the  epoch and time. Results show that the model's validation loss keep dropping instead of just finding a trivial solution very fast. This to some extent shows that the generative pre-training task is hard enough and can thus can guide the model to really capture the intrinsic structure of the graph data. It took about  12 hours for \method\ to converge. For downstream tasks, we show the convergence curve utilizing our \method\ with no-pretrain, with different data percentage. As is illustrated in Figure~\ref{fig:cv2}, \method\ can always get a more generalized model than no-pretrain, and is more robust to over-fitting since a good initialization from pre-training.

\section{Paper Title Generation Examples}
For OAG, since our attribute generation task is oriented on the paper title, we'd like to see how well our \method \ can learn to generate the title. The results are shown in table~\ref{tab:gen}. We can see that the model can capture the main meaning of each paper to be predicted, only by looking at partial neighborhoods (note that we use Attribute Generation Node for this task, which replace the input attribute as a share vector). For example, for the first sentence, our model successfully predict the key words for this paper, including `person recognition', `probabilistic', etc. This shows that the graph itself contains rich semantic information, and explains why a pre-trained model can generalize well to downstream tasks.

\end{document}